\documentclass[pdflatex,sn-mathphys-num]{sn-jnl}

\usepackage{graphicx}%
\usepackage{multirow}%
\usepackage{amsmath,amssymb,amsfonts}%
\usepackage{amsthm}%
\usepackage{mathrsfs}%
\usepackage[title]{appendix}%
\usepackage{xcolor}%
\usepackage{textcomp}%
\usepackage{manyfoot}%
\usepackage{booktabs}%
\usepackage{algorithm}%
\usepackage{algorithmicx}%
\usepackage{algpseudocode}%
\usepackage{listings}%

\usepackage[table]{xcolor}

\theoremstyle{thmstyleone}%
\theoremstyle{thmstyletwo}%

\theoremstyle{thmstylethree}%

\begin{document}

\title[Article Title]{Real-Time Crowd Counting for Embedded Systems with Lightweight Architecture}


\author[1]{\fnm{Zhiyuan} \sur{Zhao}}\email{tuzixini@gmail.com}
\equalcont{These authors contributed equally to this work.}

\author[2]{\fnm{Yubin} \sur{Wen}}\email{thebirdwyb@mail.nwpu.edu.cn}
\equalcont{These authors contributed equally to this work.}

\author[2]{\fnm{Siyu} \sur{Yang}}\email{yangsy@mail.nwpu.edu.cn}

\author[1,2]{\fnm{Lichen} \sur{Ning}}\email{ninglc@mail.nwpu.edu.cn}

\author[3]{\fnm{Yuandong} \sur{Liu}}\email{LYD2022@mail.nwpu.edu.cn}

\author*[1,2]{\fnm{Junyu} \sur{Gao}}\email{gjy3035@gmail.com}

\affil[1]{\orgdiv{Institute of Artificial Intelligence (TeleAI)}, \orgname{China Telecom}, \country{China}}

\affil[2]{\orgdiv{School of Artifcial Intelligence, Optics and Electronics (iOPEN)}, \orgname{Northwestern Polytechnical University}, \country{China}}

\affil[3]{\orgdiv{Honors College}, \orgname{Northwestern Polytechnical University}, \country{China}}


\abstract{ Crowd counting is a task of estimating the number of the crowd through images, which is extremely valuable in the fields of intelligent security, urban planning, public safety management, and so on. However, the existing counting methods have some problems in practical application on embedded systems for these fields, such as excessive model parameters, abundant complex calculations, \emph{etc}. The practical application of embedded systems requires the model to be real-time, which means that the model is fast enough. Considering the aforementioned problems, we design a super real-time model with a stem-encoder-decoder structure for crowd counting tasks, which achieves the fastest inference compared with state-of-the-arts. Firstly, large convolution kernels in the stem network are used to enlarge the receptive field, which effectively extracts detailed head information. Then, in the encoder part, we use conditional channel weighting and multi-branch local fusion block to merge multi-scale features with low computational consumption. This part is crucial to the super real-time performance of the model. Finally, the feature pyramid networks are added to the top of the encoder to alleviate its incomplete fusion problems. Experiments on three benchmarks show that our network is suitable for super real-time crowd counting on embedded systems, ensuring competitive accuracy. At the same time, the proposed network reasoning speed is the fastest. Specifically, the proposed network achieves 381.7 FPS on NVIDIA GTX 1080Ti and 71.9 FPS on NVIDIA Jetson TX1.}




\maketitle

\section{Introduction}\label{sec1}
In recent years, with the increase in population, accurately estimating the number of the crowd from pictures has become a very significant issue in intelligent surveillance. Estimating the number of crowds from images is called crowd counting, which is significant in population control, public safety management, urban planning, and intelligent transportation systems \cite{Zhou2019,Zhu2022,Wang2022,Determe2021,jiang2018deep,AIFlowReport}. Generally, most of the existing methods \cite{liu2025fast,CSRNet,UCF-QNRF} pay attention to accuracy and neglect the efficiency of models. These models have high precision and low efficiency, which does not apply to real-time surveillance application tasks. This real-time performance of the model means that it has high efficiency and rapid inference ability. Nevertheless, the limitations of the above methods impede the application of embedded devices for intelligent surveillance. 
\begin{figure}[ht]
    \centering
    \begin{minipage}{0.45\textwidth}
        \centering
        \vspace{4.4mm} 
        \includegraphics[width=\textwidth]{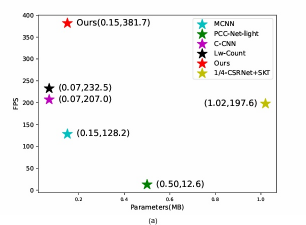}
    \end{minipage} 
    \hspace{0.05\textwidth} 
    \begin{minipage}{0.45\textwidth}
        \centering
        \includegraphics[width=\textwidth]{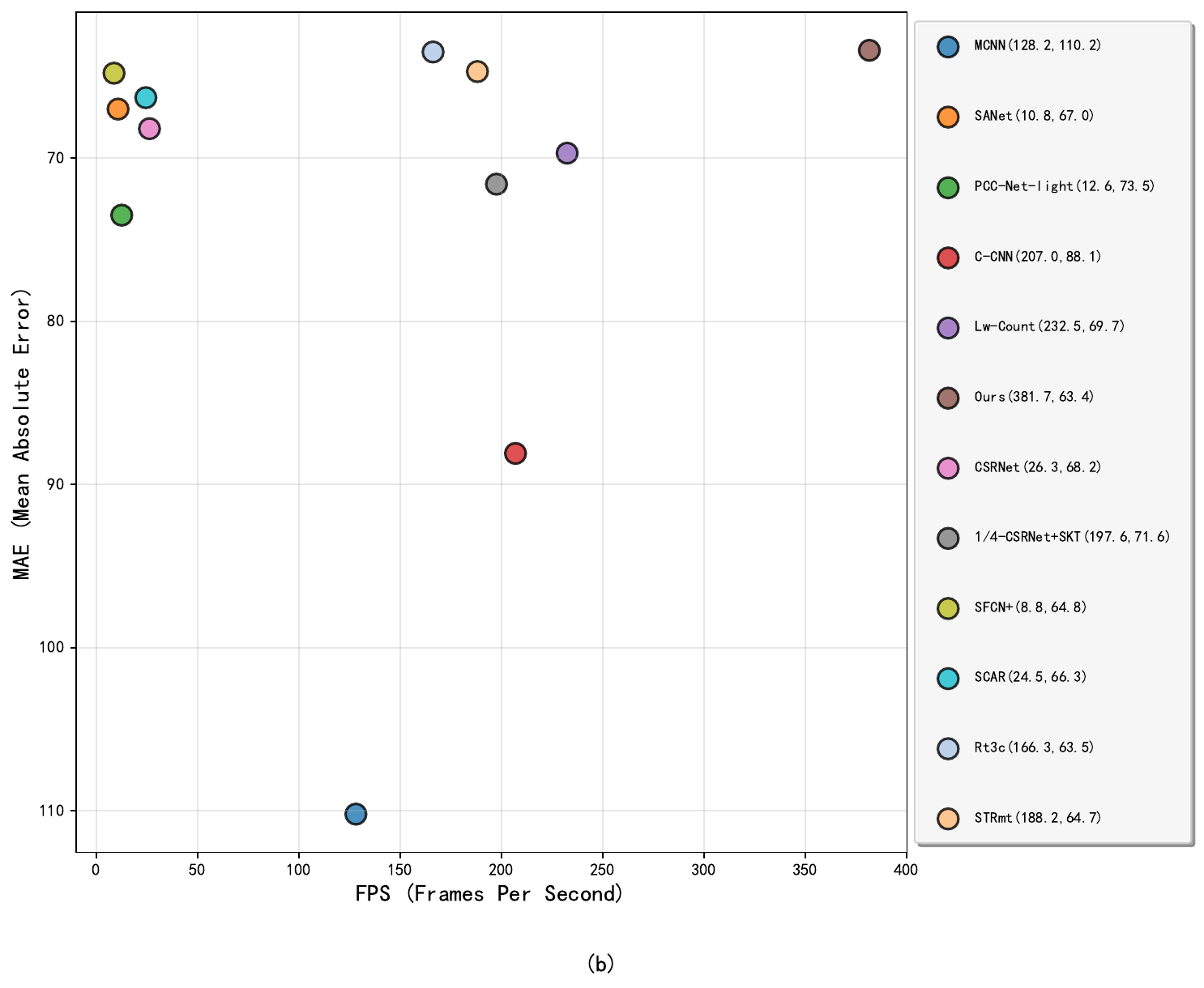}
    \end{minipage}

    \caption{
        \textbf{(a):} Comparison with other networks in parameter quantity and FPS. FPS is tested on NVIDIA GTX 1080Ti.
        \textbf{(b):} This chart shows the performance comparison of different methods in terms of frame rate (FPS) and mean absolute error (MAE) on the SHHA dataset. 'Ours' method achieves the highest frame rate (approximately 380 FPS) and the lowest mean absolute error (approximately 65 MAE), outperforming other existing methods in both speed and accuracy.
    }
    \label{fig:1}
\end{figure}

For example, the large number of parameters in the above methods needs larger memory of RAM, and the complex calculation of them has a demand on high computation power. Recently, some researchers focus on lightweight networks \cite{MobileCount,PN} with fewer parameters for crowd counting. For instance, Gao \emph{et al}. \cite{PCC-Net} proposes a lightweight perspective crowd counting network (PCC-Net-light), and Liu \emph{et al}. \cite{SKT} proposes an efficient crowd counting method (1/4-CSRNet+SKT) based on structured knowledge transfer. However, although they reduce the number of model parameters, they cannot realize the fastest inference, as shown in Fig. \ref{fig:1}. As a result, they are also not suitable for super real-time tasks.

To remedy these problems, we design a super real-time network that is oriented to low-power devices (e.g. NVIDIA Jetson TX1, NVIDIA Jetson Xavier)\cite{MiN}. The network consists of a stem network, an encoder, and a decoder. The details are as follows: 1) The stem network adopts the large convolution kernels of 9, 7, and 5. These kernels are very useful for increasing the receptive field and extracting detailed preliminary features. 2) The extracted features are divided into multi-scale branches through the down-sampling operation and fed to the encoder. In the meantime, the extracted features in multi-scale branches experience Conditional Channel Weighting (CCW) \cite{Lite-HRNet,PiNDA,PiNGDA} and Multi-branch Local Fusion (MLF) block to merge them only in down-sampling, maintaining the small scales of features in fusion and achieving the low computational consumption effectively. 3) The decoder, the Feature Pyramid Networks (FPN) \cite{FPN} are adopted to integrate the encoder's outputs of multi-scale branches, which alleviates the problem of incomplete fusion caused by the local fusion in the encoder. Experiments show that this network realizes the fastest inference and pledges accuracy, compared with state-of-the-arts.

The summary of our major contributions is as follows:

(1) We propose a novel stem–encoder–decoder framework explicitly optimized for super real-time crowd counting, achieving an optimal balance between accuracy and efficiency. The design is tailored for low-power embedded devices used in intelligent surveillance and public safety applications.

(2) We are the first to introduce the Conditional Channel Weighting (CCW) block into the crowd counting domain, enabling adaptive feature selection across resolutions. Furthermore, we design a new Multi-branch Local Fusion (MLF) block to perform efficient multi-scale feature aggregation with minimal computational cost. These two modules jointly form a lightweight yet expressive encoder that significantly enhances both accuracy and speed.

(3) We establish a new state-of-the-art in inference efficiency, reaching 381.7 FPS on NVIDIA GTX 1080Ti and 71.9 FPS on NVIDIA Jetson TX1, surpassing all existing lightweight models while maintaining competitive accuracy.

\section{Related Work}\label{section:Related}

This section consists of two parts: (A) Crowd counting. (B) Real-time or lightweight network.  

\subsection{Crowd counting}
The methods for crowd counting have three categories: (1) Detection-based methods. (2) Regression-based methods. (3) Methods based on density estimation.

\textbf{Detection-based methods.}
An early approach to addressing the challenges of crowd scene analysis \cite{group, Quantifying,zhang2018review} involved the application of detection-based algorithms \cite{detection1}, \cite{detection2}. These algorithms operate by counting individuals through the process of scanning a set of detectors over two consecutive frames within a video sequence. While this method has shown utility in certain contexts, it presents several inherent limitations, particularly in highly congested or densely packed environments. In such cases, the inter-obstruction between individuals within the crowd may significantly hinder the detector’s ability to accurately identify and track individual subject \cite{han2022dr}. This phenomenon, often referred to as crowd occlusion, leads to a deterioration in detector performance, thereby diminishing the precision and overall reliability of the counting process. As a result, the final accuracy of the model is often compromised, especially when the crowd density exceeds a threshold where individual identification becomes increasingly difficult.

\textbf{Regression-based methods.}
To address the limitations of detection-based methods, researchers have increasingly turned to regression-based approaches \cite{regress,regress1}. These methods focus on leveraging global image features, such as gradients, textures, and other holistic attributes, to estimate crowd density or count. While this approach proves useful in highly congested environments where detection becomes challenging, relying solely on low-level features can lead to the omission of key details specific to the target. This results in inaccuracies, especially in local regions of the image where precision is critical for counting tasks \cite{wan2023modeling, wan2020kernel,wan2024gene}. To further improve the robustness of these methods, recent advancements have incorporated contextual information \cite{xiong2017spatiotemporal} have developed cross-scene crowd counting techniques to enhance model generalization . These enhancements allow for better adaptation across diverse environments, helping to mitigate some of the shortcomings associated with traditional regression-based solutions \cite{lin2024fixed}, particularly in more variable or complex scenarios.

\textbf{Methods based on density estimation.}
At present, the major methods are based on density map estimation \cite{zhao2025density, lin2023optimal}. At first, Zhang \emph{et al}. \cite{MCNN} proposes a Multi-column Convolutional Neural Network (MCNN) for crowd counting. However, in the crowded scene, MCNN is difficult to train and the micro features are difficult to extract. Afterward, to deal with the problem in the chaotic background, Yi \emph{et al}. \cite{SACCN} designs a ScaleAware Crowd Counting Network (SACCN). Thereafter, some researchers gradually begin to show solicitude for improving the efficiency of the model, so they try to realize the lightweight of the network. For example, Gao \emph{et al}. \cite{PCC-Net} proposes a lightweight Perspective Crowd Counting Network (PCC-Net-light), which counts crowds from a single image. Shi \emph{et al}. \cite{C-CNN} proposes a Compact Convolutional Neural Network (C-CNN) that learns a more efficient model. Yuan \emph{et al}. \cite{yuan2025distance} introduces a framework leveraging head size variations to estimate real crowd distribution and count crowds to investigate the true distribution of crowd density in the physical world. proposes a However, although these networks ensure a low amount of model parameters, they cannot achieve the fastest inference. As a result, they do not have super real-time performance. 

Moreover, with the rapid advancement of large language models (LLMs) and their exceptional performance across diverse vision-language tasks \cite{lin2025webuibench, li2025llms,PiNI}, several recent studies have attempted to extend them to the domain of crowd counting \cite{wan2024robust}. These approaches show strong generalization in few-shot and zero-shot scenarios, reducing the need for dense annotations \cite{fan2024learning, wang2025diffusion, bai2024combating}.
However, despite these advantages, our quantitative evaluation and additional experiments reveal that current LLM-based frameworks remain far from meeting real-time requirements. For instance, representative models such as BLIP-2 (11B parameters) and LLaVA (7B parameters) require 15–22 GB of GPU memory and exhibit inference latency of 1.0–1.3 s per 224×224 image on an NVIDIA A100 GPU. When deployed on embedded platforms such as Jetson TX1, these models fail to execute a single forward pass within 10–12 seconds, consuming nearly all available memory. Even smaller-scale models with several hundred million parameters (e.g., MiniGPT-4-tiny) still incur >500 ms per frame, which is significantly slower than the 33 ms threshold (30 FPS) required for real-time applications.
Our own tests corroborate these findings, showing that none of the evaluated LLM-based models can exceed 0.1 FPS on Jetson TX1, while our proposed lightweight CNN-based network achieves over 70 FPS on the same device. These quantitative comparisons demonstrate that current LLM-based approaches are computationally prohibitive for real-time embedded crowd counting. Therefore, although LLMs are promising for few-shot or zero-shot settings, this paper focuses on designing lightweight CNN architectures capable of super real-time inference on low-power hardware.

\begin{figure*}[ht]
	\centering
	\includegraphics[width=5in,height=2in]{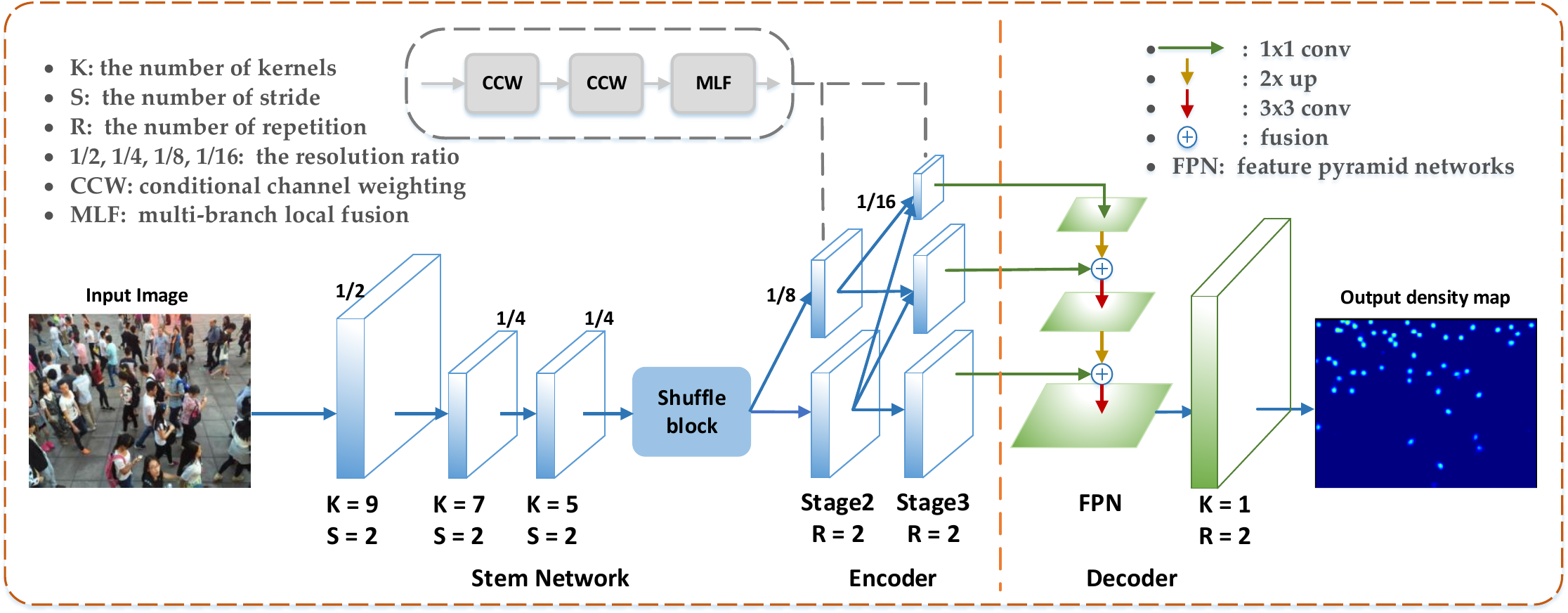}
	\caption{Our network for crowd counting tasks. The left dotted line in the figure is the stem network and encoder structure, which passes through the stem network (containing three convolutions and shuffle blocks) and two stages (containing two CCW blocks and one MLF block respectively). CCW stands for conditional channel weighting and MLF stands for multi-branch local fusion. The right dotted line is the decoder structure, and the output density map is finally obtained through feature pyramid networks (FPN) and two layers for feature regression.}
	\label{fig:2}
\end{figure*}

\subsection{Real-time or lightweight network}

Nowadays, in the field of deep learning, it has turned into a hot topic of conversation that how to use a neural network architecture to realize the best balance between accuracy and performance \cite{han2015two, tao2008bayesian, zhao2021reconstructive}. Some researchers begin to lighten the network to improve the efficiency of models. Namely, lightweight means a low number of parameters and simple floating-point operations, which is also suitable for real-time tasks. Meanwhile, for low-power devices, lightweight saves more resources and time. The followings are some networks: SqueezeNet \cite{SqueezeNet} is a small Deep Neural Networks (DNN) architecture whose parameters are reduced by 50 times to less than 0.5 MB using model compression techniques. RRTrN achieves powerful feature extraction while significantly reducing the number of parameters by leveraging recursive learning and a combination of convolutional layers and a transformer unit \cite{zhou2024rrtrn}. MobileNetV1 \cite{MobileNets} uses depth separable convolution to reduce the parameter and calculation amount, and proposes two hyperparameters Width Multiplier and Resolution Multiplier to balance the time and accuracy. ShuffleNet \cite{ShuffleNet} uses group convolution and channel shuffle to decrease the number of model parameters, achieving higher efficiency than MoblieNet. MobileNetV2 \cite{MobileNetV2} introduces the residual structure and Linear bottleneck layer, in which Linear executors are implemented by removing the ReLU after the second Conv 1$\times$1. This network structure achieves high performance. Ma \emph{et al}. proposes ShuffleNetV2 \cite{ShuffleNetV2}, which improves ShuffleNetV1 through four criteria. Based on this, we design a network to achieve fast inference time (parameter is 0.15 MB, FLOPs is 1.32 G). The network has super real-time performance and is available for intelligent surveillance and low-power devices in other fields. Ding \emph{et al}. \cite{ding2024ff} leveraging deep networks and lightweight networks to handle crucial and non-crucial data separately, the real-time performance of the network is enhanced.
Besides, some researchers \cite{zhou2025scale,li2025llms, guo2025enhancing} contribute to model pruning, lightweight optimization, and real-time performance by introducing adaptive pruning methods, efficient multimodal fusion frameworks, and robust security benchmarks. These advancements reduce computational complexity, improve inference speed, and enhance task-specific robustness, making AI models more efficient \cite{FaceModeling}and suitable for dynamic real-time applications. 
Some approaches \cite{zhang2024integrating, wang2024embedding, gao2025combining,EyeFixations} achieve this objective by incorporating efficient fine-tuning strategies, designing compact modules, and optimizing the loss function, which collectively contribute to reducing computational overhead while enhancing model accuracy.

\begin{figure}[ht]
	\centering
	\includegraphics[width=4in,height=2in]{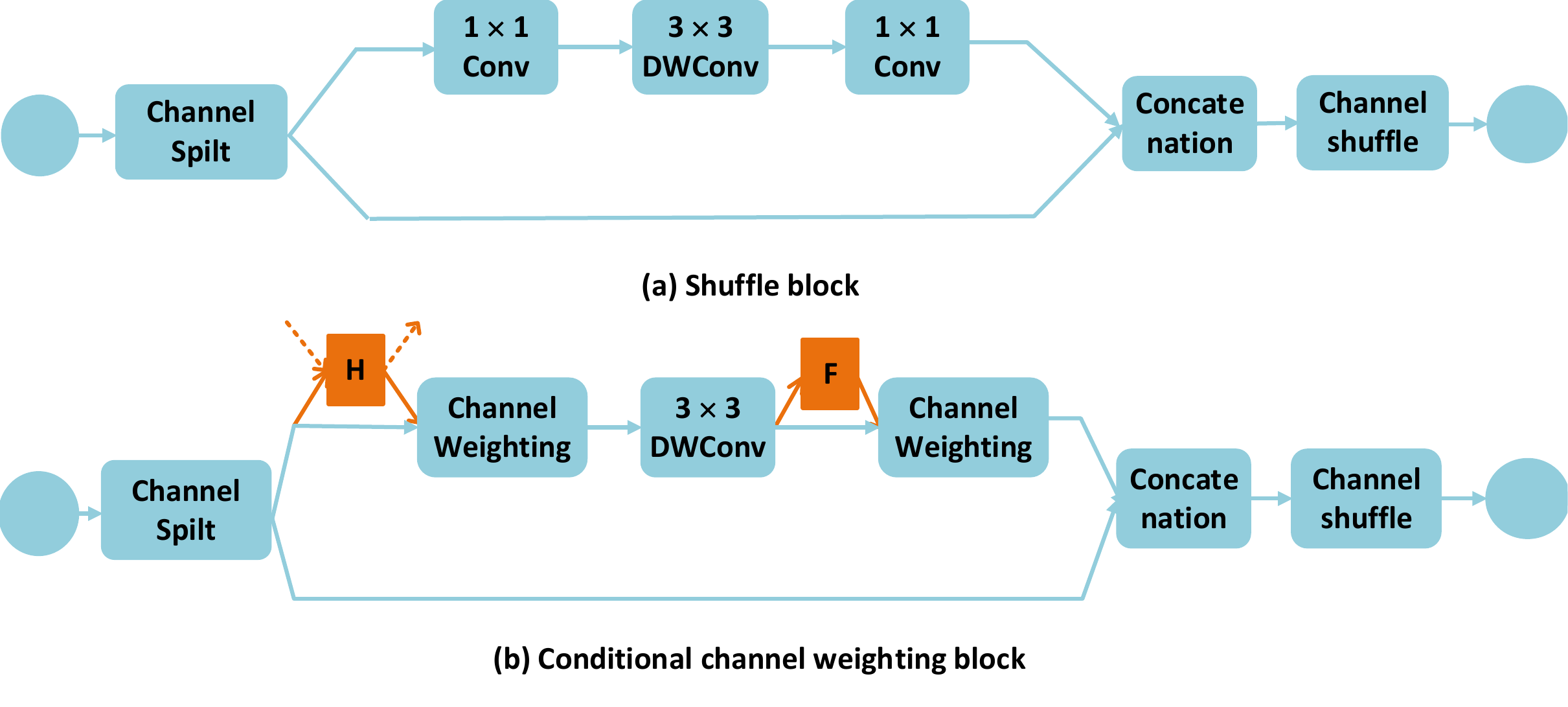}
	\caption{Two types of blocks. In (b), H in the first orange box represents the cross-resolution weight function from different branches, where the line of dashes means the weight representation assigned to other branches. In the other orange box, F represents the spatial weight function.}
	\label{fig:3}
\end{figure}

\section{Methodology} \label{sec:3}

In this section, we present the proposed super real-time model for the crowd counting task. According to Figure \ref{fig:2}, the model is an efficient stem-encoder-decoder structure. 

\subsection{Stem Network}

At this stage, for preferably extracting the features from images, the following three operations are designed, each of which plays an indispensable role, namely: (1) Early down-sampling. (2) Large convolution kernels. (3) Shuffle block.

\textbf{Early down-sampling.}
After the initial image input, the visual information in it is highly redundant in space, so it should be compressed for representation. This early down-sampling operation greatly cuts down the size of feature maps, which changes the resolution to 1/4 of the original image through a series of convolution operations. 

\textbf{Large convolution kernels.}
In the stem network, to obtain more detailed head features, the large convolution kernels of 9, 7, and 5 are used, which commendably expand the receptive field\cite{VPN}. The ablation study in Section \ref{section:Experiments} also shows that large kernels are the most effective compared to the small kernels and the dilated kernels, and the additional amount of calculation is worth while pursuing accuracy.

\textbf{Shuffle block.}
Shuffle block is proposed in ShuffleNetV2 \cite{ShuffleNetV2}. According to Figure \ref{fig:3} (a), it divides the information channel into two parts and through convolution, concatenating and shuffling to achieve the effect of mixed information.
\begin{figure*}[ht]
	\centering
	\includegraphics[width=5.4in,height=2.8in]{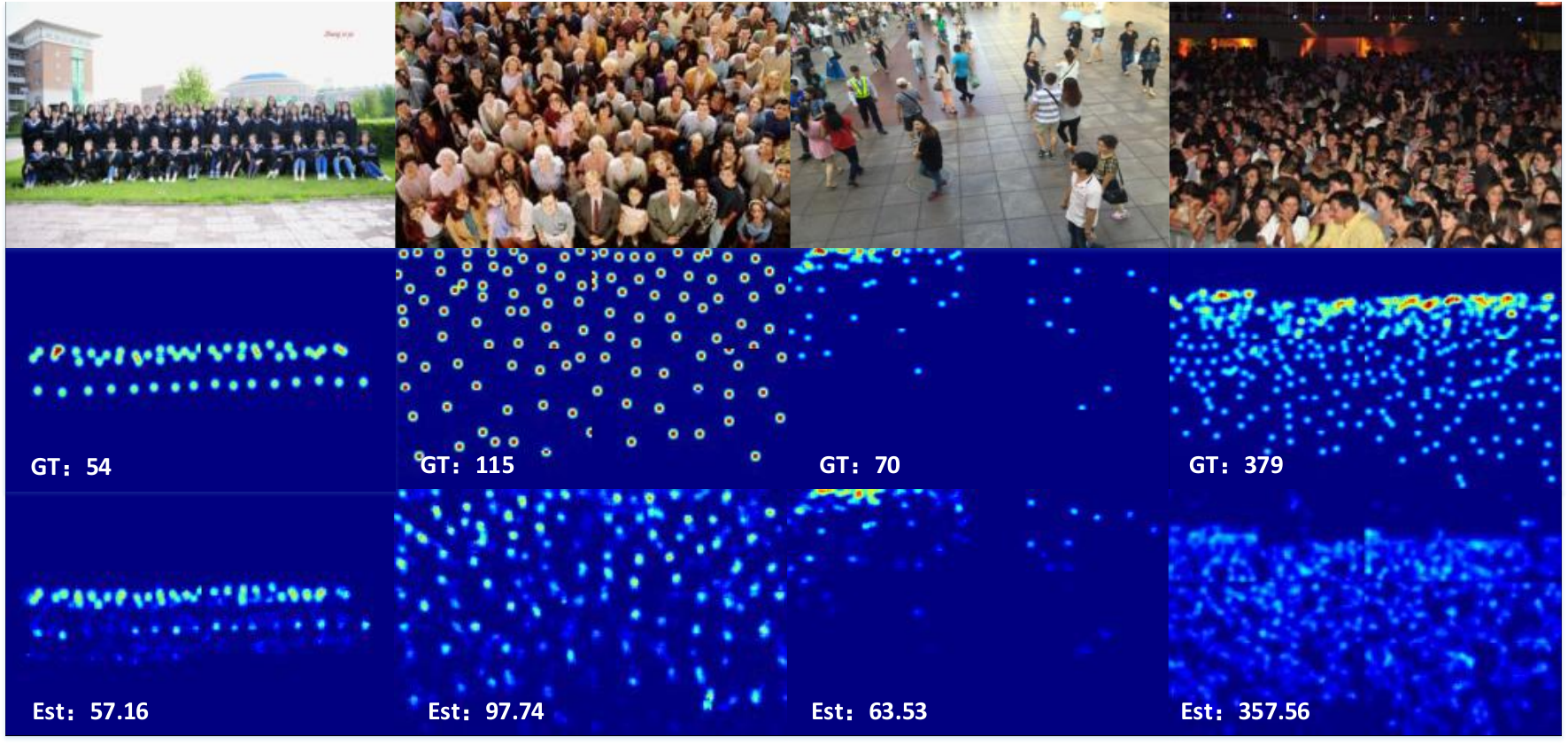}
	\caption{The first row displays images on NWPU-Crowd, ShanghaiTech, and UCF-QNRF test sets. The second row lists the corresponding ground truth density maps (GT). The third row lists predicted density maps (Est). Numbers represent the actual and predicted number of people, respectively.}
	\label{fig:4}
\end{figure*}

\subsection{Encoder Architecture}
Two stages are meticulous in design for the encoder part to decrease the quantity of parameters and floating-point operations. After the stem network, the extracted features are divided into multi-scale branches by down-sampling into these two stages, in which the minimum resolution is $\frac{1}{16}$W$\times$$\frac{1}{16}$H. These two stages include a series of components, and each component has two Conditional Channel Weighting (CCW) blocks and one Multi-branch Local Fusion (MLF) block, which repeats the component twice for each stage. The role of CCW and MLF is demonstrated in the ablation study of Section \mbox{\ref{section:Experiments}}.

\textbf{Conditional channel weighting.}
This block is used to calculate the channel weights of single and multi resolution. CCW is proposed in Lite-HRNet \cite{Lite-HRNet}, as shown in Figure \ref{fig:3} (b). Similar to the shuffle block, the convolution operation is replaced by the element-by-element weighting operation through the cross resolution weight and spatial weight function to reduce complexity.

\textbf{Multi-branch local fusion.}
This block is meticulous in design by us to fuse the multi-scale features after the CCW block, which are fused by down-sampling and sum operation. Specifically, in MLF block, this is divided into two parts: 1) For layers with the same resolution that of which number and size of channels are the same, features are used directly. 2) For layers with different resolutions, channel conversion and down-sampling are carried out from high to low resolution (through 3$\times$3 Conv with stride = 2 in series). As shown in Figure \ref{fig:2}, the maximum number of branches is 3, their resolutions are $\frac{1}{4}$, $\frac{1}{8}$, $\frac{1}{16}$ of the original images, the corresponding amount of channels are 36, 64, 96 separately. 
\subsection{Decoder Architecture}

In this part, Feature Pyramid Networks (FPN) \cite{FPN} is used to alleviate the problem of incomplete multi-branch fusion left by the encoder, and then two layers for feature regression are carried out to output the final feature map. 

\textbf{Feature pyramid networks.}
In this work, the encoder is not fully connected, in which the low-resolution channel information after down-sampling is not transmitted to the upper layer. As a result, for alleviating the problem of incomplete integration originating from the local fusion in the encoder, FPN is used as a decoder to combine multi-scale features to deal with the size of different images. The decoding process of FPN is carried out by up-sampling and lateral connection. Specifically, firstly, the low-resolution feature map of the encoder is sampled twice, and then the result is fused with the feature map of the same size generated by convolution of 1$\times$1, so that the details can be located. After the fusion, the convolution of 3$\times$3 is used to check the convolution of each fusion result to eliminate the aliasing effect of up-sampling. At the same time, this process is iterative until the final density map is created. The parameter quantity of FPN is 0.085MB.

\textbf{Two layers for feature regression.}
After the FPN decoding operation, each layer has an output feature map. The three feature maps are combined, and then the feature regression is carried out through two 1$\times$1 convolution operations to make the amount of channels 1 and output a final prediction map.

\subsection{Loss Functions}
We convert the inputs into a density map and use the mean square error loss. The formula is as follows:
\begin{equation}\label{key}
	Loss = \frac{1}{N}\sum_{j=1}^{N}||D_{j}^{P}-D_{j}^{G}||_{2}^{2},
\end{equation}
where $Loss$ is the mean square error loss, $N$ stands for the amount of input images, $D_{j}^{P}$ stands for predicted density maps, and $D_{j}^{G}$ means ground truth maps, $j$ means the $j$-th input image, $||\cdot ||_{2}^{2}$ expresses the Euclidean metric.

\begin{sidewaystable}
\caption{Tables which are too long to fit, should be written using the ``sidewaystable'' environment as shown here}\label{tab3}

\label{table:1}
	\begin{tabular}{c c c c c c c c c c c c c c c} 
		\toprule 
		&\multicolumn{3}{c}{NWPU-Crowd}&\multicolumn{2}{c}{SHHA}&\multicolumn{2}{c}{SHHB}&\multicolumn{2}{c}{UCF-QNRF}  &  &  &\\
		Method  & MAE & MSE & NAE & MAE & MSE & MAE & MSE & MAE & MSE&FLOPs & Params  & Time &FPS & AES \\
		\midrule 
		MCNN\cite{MCNN} & 232.5& 714.6  & 1.063& 110.2 & 173.2 & 26.4 & 41.3& 277 & 426 & 11.99 & 0.15 &7.8& 128.2 & 12.87\\
		SANet\cite{SANet} & 190.6 & 491.4 & 0.991 &67.0& 104.5 &	8.4 & 13.6 &-&-&40.10& 1.39 & 93.0&10.8 & 4.15\\
		
		CMTL\cite{CMTL} &-&- &- & 101.3  & 152.4 & 20.0 & 31.1 & 252& 514 &- & 2.36 &- &-  & -\\
		
		ACSCP\cite{ACSCP} &- &- &-&75.7 &102.7&17.2  & 27.4 &-  & - &- &5.10 &- &-  & - \\
		
		PCC-Net-light\cite{PCC-Net}&167.4 & 566.2& 0.444& 73.5  &  124.0 & 11.0  & 19.0&148.7 &247.3  &72.797 &0.50&79.3&12.6 & 4.85\\
		
		TDF-CNN\cite{TDF-CNN}&-& -&- &97.5 & 145.1 &  20.7 &  32.8 & - & - & -& 0.13& - & - & -\\
		
		C-CNN\cite{C-CNN} &- &- & -&88.1  & 141.7 & 14.9 &  22.1 & - & -&- &\underline{\textbf{0.07}} & 4.8&207.0 & - \\
		CACrowdGAN\cite{Zhu2022} &- &- & -&\underline{\textbf{61.8}}&\underline{\textbf{98.8}}&\underline{\textbf{7.2}}&  12.9 & 109.5 &173.1 &-&37.94&-&-  & - \\
		Lw-Count\cite{Liu2022} &- &- & -&69.7&\underline{100.5}&10.1& 12.4&149.7&238.4&-&\underline{\textbf{0.07}}& 4.3&232.5& -\\
        
		\cellcolor{lightgray!30} Ours& \cellcolor{lightgray!30} \cellcolor{lightgray!30} \underline{\textbf{102.5}}&\cellcolor{lightgray!30} \underline{ 388.6}  \cellcolor{lightgray!30} &\cellcolor{lightgray!30} \underline{0.265} & \cellcolor{lightgray!30} \underline{63.4}  &\cellcolor{lightgray!30} 106.3	&\cellcolor{lightgray!30} \underline{7.3}&\cellcolor{lightgray!30} \underline{\textbf{11.2}} &\cellcolor{lightgray!30} \underline{104.1} &\cellcolor{lightgray!30} \underline{172.2}&\cellcolor{lightgray!30} \underline{\textbf{1.32}}& \cellcolor{lightgray!30} 0.15 &\cellcolor{lightgray!30} \underline{\textbf{2.6}}& \cellcolor{lightgray!30} \underline{\textbf{381.7}}
        &\cellcolor{lightgray!30} \underline{\textbf{46.58}}\\
		\midrule 
		CSRNet\cite{CSRNet} & 121.3 & \underline{\textbf{387.8}} & 0.604 & 68.2 & 115.0 & 10.6 &16.0 &- & -&183.00& 16.26  &38.0 & 26.3 & 3.04\\
		1/4-CSRNet+SKT\cite{SKT}&- &-&-&71.6&114.4&{7.5}&{11.7}&144.4&234.6 &\underline{11.64}& \underline{1.02}&\underline{5.1} &\underline{197.6} & -\\
		C3F-VGG\cite{C3F-VGG} & 127.0&439.6  &0.411  &- &- &- &-  &- &- &123.81& 7.34 &21.0& 47.6 & 3.10\\
		SFCN$\dagger$\cite{SFCN+}& \underline{{105.7}}&424.1  & \underline{\textbf{0.254}}&{64.8}&{107.5}&7.6&13.0&\underline{\textbf{102.0}}&\underline{\textbf{171.4}}&272.65& 36.81  &114.0&8.8 & 3.02\\
		SCAR\cite{C3F-VGG}&110.0&495.3 &0.288  &66.3 &114.1 &9.5 & 15.2 &- &- &182.86& 16.29&40.8&  24.5 & 3.03\\
        
        Rt3c\cite{wang2024rt3c}&108.2 &417.2 &0.292  &63.5 &102.2 &7.5 & \underline{11.3} &- &- &12.49& 5.45 &16.3&  166.3 & 6.54\\
        
        STRmt\cite{sun2024strmt}&120.1&435.8 &0.321  &64.7 &111.8 &7.6 & 11.8 &- &- &2.04& 0.24 &8.2&  188.2 & 30.71\\
        
		\bottomrule
	\end{tabular}
    \label{main_tabel}
\end{sidewaystable}

\section{Experiments}\label{section:Experiments}

This section has five following parts: (A) It introduces three indicators: MAE, MSE, and NAE. (B) It introduces three datasets. (C) It introduces the comparison of experimental results with other existing methods on these three datasets. (D) It introduces three ablation studies. (E) It reports the inference speed results on different system modules.


\subsection{Metrics}
In experiments, three metrics are used to assess the function of the model, which are Mean Absolute Error (MAE), Mean Square Error (MSE), and Mean Normalized Absolute Error (NAE). They are defined as follows:
\begin{equation}\label{key}
	MAE = \frac{1}{N}\sum_{i=1}^{N}|y_{i}^{P}-y_{i}^{G}|,
\end{equation}

\begin{equation}\label{key}
	MSE =\sqrt{\frac{1}{N}\sum_{i=1}^{N}|y_{i}^{P}-y_{i}^{G}|^{2}},
\end{equation}

\begin{equation}\label{key}
	NAE = \frac{1}{N}\sum_{i=1}^{N}\frac{|y_{i}^{P}-y_{i}^{G}|}{y_{i}^{P}},
\end{equation}
where $N$ is the quantity of images in these three datasets. $y_{i}^{G}$ is the actual quantity and $y_{i}^{P}$ is the predicted number of the $i$-th test image. Moreover, MAE is a metric to assess the accuracy of calculated crowd numbers, while MSE displays the robustness of estimated crowd numbers.

\subsection{Implementation Details}
To ensure the reproducibility of our experiments, we provide detailed information on the training configuration and hyperparameter settings as follows:
\begin{itemize}
\item  \textbf{Framework}. All experiments were implemented using PyTorch 1.6.

\item \textbf{Optimizer}. The model was trained using the Adam optimizer, with parameters $\beta_1 = 0.9$ and $\beta_2 = 0.999$.

\item \textbf{Learning rate and schedule}. The initial learning rate was set to $1e-4$ and decayed following a cosine annealing schedule with a minimum learning rate of $1e-6$.

\item \textbf{Batch size and epochs}. The batch size was 16, and training was conducted for 300 epochs on each dataset.

\item \textbf{Weight decay}. A weight decay of $5e-4$ was applied to regularize the network parameters and prevent overfitting.

\item \textbf{Data preprocessing}. All images were resized to $512×1024$ pixels and normalized to zero mean and unit variance. Random horizontal flipping and random cropping were employed for data augmentation during training.

\item \textbf{Loss function}. The mean square error (MSE) loss, defined in Eq. (2), was adopted for density map regression.

\item \textbf{Initialization}. The network weights were initialized using He normal initialization, and all convolutional layers employed ReLU activations unless otherwise specified.

\item \textbf{Evaluation}. During testing, no data augmentation was applied. All models were evaluated using MAE, MSE, and NAE metrics as described in Section 4.1.
\end{itemize}

\subsection{Datasets}


In this section, it introduces three datasets: UCF-QNRF \cite{UCF-QNRF}, NWPU-Crowd \cite{NWPU-Crowd}, ShanghaiTech \cite{MCNN}. Among them, ShanghaiTech is a small-scale dataset. Since small-scale datasets are easy to be over-fitted, it also conducts experiments on the two large-scale datasets (UCF-QNRF, NWPU-Crowd), which evaluate a network more effectively and accurately.

\textbf{UCF-QNRF.}
This dataset contains 1,535 dense crowd images, of which 1,201, 334 images make use of training and testing respectively. These images include the most diverse viewpoint set, illumination change, and density, making this dataset more realistic and difficult.

\textbf{NWPU-Crowd.}
This dataset includes 5,109 images, of which 3,109, 500, and 2,000 images are used for training, validation, and testing respectively. There are 2,133,375 annotation headers with points and boxes. This is the latest and largest dataset at present, which contains a variety of lighting scenes with the largest density range (0$\sim$20,033).

\textbf{ShanghaiTech.}
This dataset consists of two parts: SHHA and SHHB. SHHA is 482 images captured from the Internet at random. Among them, 182 make use of testing sets, the rest are training sets. SHHB is 716 images shot from the bustling commercial street. Among them, 316 make use of testing sets, the rest are training sets.

\subsection{Results}

We use the density estimation method to test the network on NWPU-Crowd \cite{NWPU-Crowd}, ShanghaiTech \cite{MCNN}, and UCF-QNRF \cite{UCF-QNRF} datasets to obtain their respective density maps. The visualization results are revealed in Figure \ref{fig:4}. Actually, comparing the ground truth with the prediction map, it is crystal clear that the network has high accuracy.

While several baselines such as CACrowdGAN achieve slightly lower MAE on specific datasets, their computational cost and inference latency are significantly higher. For example, CACrowdGAN and SFCN+ require hundreds of millions of parameters and achieve less than 50 FPS on the same hardware, making them impractical for real-time embedded applications. In contrast, our network maintains competitive accuracy with an extremely compact structure (0.15 MB) and minimal computational demand (1.32 GFLOPs), achieving 381.7 FPS on GTX 1080Ti and 71.9 FPS on Jetson TX1.

This phenomenon reveals a crucial insight: accuracy alone cannot reflect a model’s real-world usability in embedded systems. Our design deliberately sacrifices marginal accuracy (approximately 2–3 MAE difference) to gain more than 5× to 10× acceleration in inference speed. This balance, quantitatively measured by our proposed Accuracy–Efficiency Score (AES), demonstrates that our model delivers the best trade-off between accuracy and efficiency among all compared methods.

Furthermore, we observe that methods with deep backbones (e.g., CSRNet) tend to overfit small-scale datasets such as SHHA but degrade sharply on large-scale benchmarks like NWPU-Crowd, whereas our lightweight encoder generalizes better across diverse densities. This generalization stems from the multi-branch local fusion design, which effectively integrates spatial and contextual cues without excessive parameters.

It is precisely because of the network architecture introduced in Section \ref{sec:3} that such good experimental results can be achieved. However, our network can also be optimized. In the future, we tend to squeeze the network to use a more lightweight decoder instead of FPN, so that they can be used on more marginalized devices.

To further balance accuracy and efficiency, we propose the Accuracy–Efficiency Score (AES), define as:
\begin{equation}
    AES = \omega_m \frac{1}{1-e^{0.01\cdot  MSE}} + \omega_p \frac{1}{1-e^{params}}+\omega_f \frac{1}{1-e^{0.02 \cdot FLOPs}} ,
\end{equation}

where $\omega_m, \omega_p$ and $\omega_f$ represents the weight of MSE, Params, and FPS respectively. This metric jointly considers inference speed and prediction accuracy, which are both critical for real-time crowd counting on embedded devices. As shown in Table \ref{main_tabel}, our method achieves the highest AES across benchmarks, demonstrating that it provides the best overall trade-off between accuracy and efficiency among the compared methods.

\subsection{Ablation Study}

It performs ablation experiments on the UCF-QNRF and NWPU-Crowd datasets, and reports outcomes on its validation set to prove the effectiveness of this model. The UCF-QNRF and NWPU-Crowd dataset are large datasets, covering a wider range of types, which is appropriate for evaluating the performance of the network. 

\textbf{Comparison of convolution kernels of different sizes.}

We compare convolution kernels of different sizes in the stem network. Large kernels with 9, 7, 5, small kernels with 3, 3, 3, and dilated kernels with 3, 3, 3 are used in stem tests separately. Apparently, small kernel convolution extracts limited information, which limits the accuracy that the model achieves. The proposed model aims to improve efficiency without using any pre-trained loading data, so it is necessary to use large kernel convolution to expand the receptive field to extract more effective information. Although the dilated kernels similarly expand the receptive field, it can be seen from Table \ref{table:2} that the network with large kernels achieves better results, with lower MAE, MSE, and runtime. We speculate that this is because the middle portion of the dilated convolution ignores too small head information in dense scenes. 
\renewcommand\arraystretch{2}
\begin{table}[h!]\small
	\centering
	\caption{Comparison of convolution kernels of different sizes on the UCF-QNRF validation set (resolution is 512$\times$1024) and the runtime of images is measured on RTX 3090.}
	\label{table:2}
	\begin{tabular}{c c c c c }
		\hline
		Method & Runtime(ms)& Params (MB) & MAE & MSE    \\
		\hline\hline
		Large kernels&3.45 &0.149 & 104.1&172.2  \\
		Small kernels&3.57 &0.126 & 114.2& 195.3  \\
		
		Dilated kernels&3.78 &0.126 &110.1 & 185.1  \\
		\hline
	\end{tabular}
\end{table}

%


\textbf{Test the effects of different blocks and parts.}

We test the effects of different blocks and parts. From Table \ref{table:4}, it can be seen that the overall decision-making is correct and each module and part is indispensable. A single block (CCW/MLF) is not as accurate and efficient as a complete block (CCW+MLF). Although the parameter quantity and FLOPs of the single stem part are relatively low, the accuracy is extremely poor. Although the accuracy of the increased encoder portion is only slightly lower than the complete framework, its computational complexity is high. This can be attributed to an increase in the number of mixed channels and parameters in the final regression layer due to the lack of encoder components.

\renewcommand\arraystretch{2}
\begin{table}[h!]\small
\centering
\caption{Test the effects of different blocks and parts on the UCF-QNRF validation set (resolution is 512$\times$1024).}
\label{table:4}
\begin{tabular}{c c c c c }
	\hline
	Method & FLOPs(G)& Params (MB) & MAE & MSE    \\
	\hline\hline
	CCW&1.49 &0.136 & 110.1&180.7 \\
	MLF&1.47 &0.120 & 112.1& 181.1 \\
	Stem& 1.26&0.032 &237.1&333.9 \\
	
	Stem + Encoder&1.41&0.117 &104.9 &173.6 \\
	Ours&1.32 &0.149 & 104.1&172.2 \\
	\hline
\end{tabular}
\end{table}

\textbf{Comparison of different numbers of modules.}

We compare networks that apply different numbers of modules without changing other settings. The number of module is expressed as num=(2,2), num=(2,4), num=(2,6). This means that module is repeated twice at Stage2. Meanwhile, it is repeated 2 times, 4 times, and 6 times at Stage3 respectively. The more repetitions, the higher the complexity of the network and the larger the quantity of parameters. On the contrary, the accuracy of the model may be better, and the extraction of head features is more detailed, as shown in Table \ref{table:5}.

From these results, we observe that larger convolution kernels substantially improve receptive field, enabling more accurate localization of small heads in highly congested areas. In contrast, dilated convolutions lead to gridding artifacts that degrade fine feature extraction, explaining their weaker performance. Furthermore, although our model excels in speed, we note that it occasionally underestimates counts in extremely dense regions where head boundaries overlap severely. This limitation suggests potential directions for future research, such as combining our lightweight model with density-aware loss functions to enhance robustness in ultra-dense scenes

\renewcommand\arraystretch{2}
\begin{table}[h!]\small
\centering
\caption{Comparison of different numbers of modules on the NWPU-Crowd validation set.}
\label{table:5}
\begin{tabular}{c c c c c c}
	\hline
	Method & FLOPs(G)& Params (MB) & MAE & MSE & NAE   \\
	\hline\hline
	Num=(2,2)&1.32& 0.149&102.0 & 503.9&	0.421  \\
	Num=(2,4)&1.36 &0.197 &95.3&497.6&0.352  \\
	Num=(2,6)&1.39&0.246 & 94.6 &589.8& 0.262  \\
	\hline
\end{tabular}
\end{table}
\renewcommand\arraystretch{2}
\begin{table*}[!t]\small
	\centering
	\caption{Compare the runtime and FPS of images (NWPU-Crowd test set) with different resolutions on four devices.}
\label{table:3}
\begin{tabular}{p{2cm}p{0.8cm}p{1cm}p{0.8cm}p{1cm}p{1cm}p{0.8cm}p{0.8cm}p{1cm}} 
	\hline
	&\multicolumn{2}{c}{GTX 1080Ti}  &\multicolumn{2}{c}{RTX 3090} &\multicolumn{2}{c}{NVIDIA TX1} &\multicolumn{2}{c}{NVIDIA  Xavier}\\
	\hline\hline
	Resolution  & Time&FPS &Time  &FPS& Time  &FPS & Time  &FPS \\
	\hline
	576$\times$768 & 2.62& 381.68& 2.42&413.22 &13.90 &71.94 &9.34 &107.07\\
	
	1024$\times$1024 &8.88& 112.61 & 5.79&172.71 &36.80 &27.17 &24.34 &41.08 \\
	\hline
\end{tabular}
\end{table*}
\subsection{Runtime}

The purpose of this work is to realize the super real-time crowd counting task, so the key goal is to achieve fast inference while ensuring the efficiency of this model. We compare networks with different input sizes on four hardware facilities to test and verify the super real-time function of this method. In Table \ref{table:3}, it calculates the test code on GTX 1080Ti, RTX 3090, NVIDIA TX1 and NVIDIA Xavier respectively. The resolution of the input image is 576$\times$768 and 1024$\times$1024, which is completed on the test set of NWPU-Crowd.

As seen from Table \ref{table:3}, it is obvious that this network has super real-time performance. We compared the runtime with different resolutions. Through the test of GTX 1080Ti, RTX 3090, NVIDIA TX1, and NVIDIA Xavier hardware facilities, this network achieves fast inference, and the runtime is no more than 15ms (resolution is 576$\times$768). It can be seen that this network is suitable to be applied to super real-time tasks while ensuring computational accuracy. In practical application scenarios, it better implements fast reasoning and is applicable to various devices for security early warning.

\section{Conclusion}\label{section:Conclusion}

In this paper, we propose a novel super real-time crowd counting network based on the stem-encoder-decoder framework.  This network is specifically engineered to address the challenges of efficient crowd counting in real-time applications.  Through careful architectural design and optimization, the proposed network achieves superior performance in both computational efficiency and memory utilization. 
These optimizations enable rapid and efficient inference, which is essential for real-time deployment across a range of scenarios.
Despite these optimizations, the network maintains competitive accuracy in crowd counting tasks, ensuring that precision is not sacrificed.  The effectiveness of the proposed network is validated through comprehensive experiments conducted on three benchmark datasets commonly used in the field.  The experimental results demonstrate that our network not only achieves super real-time performance but also surpasses existing methods in terms of inference speed, making it highly suitable for deployment on embedded devices with constrained computational resources.

\section*{Abbreviations}
\begin{itemize}
\item CCW: Conditional Channel Weighting
\item MLF: Multi-branch Local Fusion
\item FPN: Feature Pyramid Networks
\item DNN: Deep Neural Networks
\item CNN: Convolutional neural network
\item MAE: Mean Absolute Error
\item RMSE: Mean Square Error 
\item NAE: Mean Normalized Absolute Error (NAE)
\item FPS: Flames Per Second
\end{itemize}


\bibliography{sn-article}


\begin{thebibliography}{72}
\ifx \bisbn   \undefined \def \bisbn  #1{ISBN #1}\fi
\ifx \binits  \undefined \def \binits#1{#1}\fi
\ifx \bauthor  \undefined \def \bauthor#1{#1}\fi
\ifx \batitle  \undefined \def \batitle#1{#1}\fi
\ifx \bjtitle  \undefined \def \bjtitle#1{#1}\fi
\ifx \bvolume  \undefined \def \bvolume#1{\textbf{#1}}\fi
\ifx \byear  \undefined \def \byear#1{#1}\fi
\ifx \bissue  \undefined \def \bissue#1{#1}\fi
\ifx \bfpage  \undefined \def \bfpage#1{#1}\fi
\ifx \blpage  \undefined \def \blpage #1{#1}\fi
\ifx \burl  \undefined \def \burl#1{\textsf{#1}}\fi
\ifx \doiurl  \undefined \def \doiurl#1{\url{https://doi.org/#1}}\fi
\ifx \betal  \undefined \def \betal{\textit{et al.}}\fi
\ifx \binstitute  \undefined \def \binstitute#1{#1}\fi
\ifx \binstitutionaled  \undefined \def \binstitutionaled#1{#1}\fi
\ifx \bctitle  \undefined \def \bctitle#1{#1}\fi
\ifx \beditor  \undefined \def \beditor#1{#1}\fi
\ifx \bpublisher  \undefined \def \bpublisher#1{#1}\fi
\ifx \bbtitle  \undefined \def \bbtitle#1{#1}\fi
\ifx \bedition  \undefined \def \bedition#1{#1}\fi
\ifx \bseriesno  \undefined \def \bseriesno#1{#1}\fi
\ifx \blocation  \undefined \def \blocation#1{#1}\fi
\ifx \bsertitle  \undefined \def \bsertitle#1{#1}\fi
\ifx \bsnm \undefined \def \bsnm#1{#1}\fi
\ifx \bsuffix \undefined \def \bsuffix#1{#1}\fi
\ifx \bparticle \undefined \def \bparticle#1{#1}\fi
\ifx \barticle \undefined \def \barticle#1{#1}\fi
\bibcommenthead
\ifx \bconfdate \undefined \def \bconfdate #1{#1}\fi
\ifx \botherref \undefined \def \botherref #1{#1}\fi
\ifx \url \undefined \def \url#1{\textsf{#1}}\fi
\ifx \bchapter \undefined \def \bchapter#1{#1}\fi
\ifx \bbook \undefined \def \bbook#1{#1}\fi
\ifx \bcomment \undefined \def \bcomment#1{#1}\fi
\ifx \oauthor \undefined \def \oauthor#1{#1}\fi
\ifx \citeauthoryear \undefined \def \citeauthoryear#1{#1}\fi
\ifx \endbibitem  \undefined \def \endbibitem {}\fi
\ifx \bconflocation  \undefined \def \bconflocation#1{#1}\fi
\ifx \arxivurl  \undefined \def \arxivurl#1{\textsf{#1}}\fi
\csname PreBibitemsHook\endcsname

\bibitem[\protect\citeauthoryear{Zhou et~al.}{2019}]{Zhou2019}
\begin{barticle}
\bauthor{\bsnm{Zhou}, \binits{Q.}},
\bauthor{\bsnm{Zhang}, \binits{J.}},
\bauthor{\bsnm{Che}, \binits{L.}},
\bauthor{\bsnm{Shan}, \binits{H.}},
\bauthor{\bsnm{Wang}, \binits{J.Z.}}:
\batitle{Crowd {C}ounting {W}ith {L}imited {L}abeling {T}hrough {S}ubmodular {F}rame {S}election}.
\bjtitle{IEEE Transactions on Intelligent Transportation Systems}
\bvolume{20}(\bissue{5}),
\bfpage{1728}--\blpage{1738}
(\byear{2019})
\end{barticle}
\endbibitem

\bibitem[\protect\citeauthoryear{Zhu et~al.}{2022}]{Zhu2022}
\begin{barticle}
\bauthor{\bsnm{Zhu}, \binits{A.}},
\bauthor{\bsnm{Zheng}, \binits{Z.}},
\bauthor{\bsnm{Huang}, \binits{Y.}},
\bauthor{\bsnm{Wang}, \binits{T.}},
\bauthor{\bsnm{Jin}, \binits{J.}},
\bauthor{\bsnm{Hu}, \binits{F.}},
\bauthor{\bsnm{Hua}, \binits{G.}},
\bauthor{\bsnm{Snoussi}, \binits{H.}}:
\batitle{C{AC}rowd{GAN}: {C}ascaded {A}ttentional {G}enerative {A}dversarial {N}etwork for {C}rowd {C}ounting}.
\bjtitle{IEEE Transactions on Intelligent Transportation Systems}
\bvolume{23}(\bissue{7}),
\bfpage{8090}--\blpage{8102}
(\byear{2022})
\end{barticle}
\endbibitem

\bibitem[\protect\citeauthoryear{Wang and Breckon}{2022}]{Wang2022}
\begin{barticle}
\bauthor{\bsnm{Wang}, \binits{Q.}},
\bauthor{\bsnm{Breckon}, \binits{T.P.}}:
\batitle{Crowd {C}ounting via {S}egmentation {G}uided {A}ttention {N}etworks and {C}urriculum {L}oss}.
\bjtitle{IEEE Transactions on Intelligent Transportation Systems}
\bvolume{23}(\bissue{9}),
\bfpage{15233}--\blpage{15243}
(\byear{2022})
\end{barticle}
\endbibitem

\bibitem[\protect\citeauthoryear{Determe et~al.}{2021}]{Determe2021}
\begin{barticle}
\bauthor{\bsnm{Determe}, \binits{J.-F.}},
\bauthor{\bsnm{Singh}, \binits{U.}},
\bauthor{\bsnm{Horlin}, \binits{F.}},
\bauthor{\bsnm{De~Doncker}, \binits{P.}}:
\batitle{Forecasting {C}rowd {C}ounts {W}ith {W}i-{F}i {S}ystems: {U}nivariate, {N}on-{S}easonal {M}odels}.
\bjtitle{IEEE Transactions on Intelligent Transportation Systems}
\bvolume{22}(\bissue{10}),
\bfpage{6407}--\blpage{6419}
(\byear{2021})
\end{barticle}
\endbibitem

\bibitem[\protect\citeauthoryear{Jiang et~al.}{2018}]{jiang2018deep}
\begin{barticle}
\bauthor{\bsnm{Jiang}, \binits{X.}},
\bauthor{\bsnm{Pang}, \binits{Y.}},
\bauthor{\bsnm{Li}, \binits{X.}},
\bauthor{\bsnm{Pan}, \binits{J.}},
\bauthor{\bsnm{Xie}, \binits{Y.}}:
\batitle{Deep neural networks with elastic rectified linear units for object recognition}.
\bjtitle{Neurocomputing}
\bvolume{275},
\bfpage{1132}--\blpage{1139}
(\byear{2018})
\end{barticle}
\endbibitem

\bibitem[\protect\citeauthoryear{An et~al.}{2025}]{AIFlowReport}
\begin{botherref}
\oauthor{\bsnm{An}, \binits{H.}},
\oauthor{\bsnm{Hu}, \binits{W.}},
\oauthor{\bsnm{Huang}, \binits{S.}},
\oauthor{\bsnm{Huang}, \binits{S.}},
\oauthor{\bsnm{Li}, \binits{R.}},
\oauthor{\bsnm{Liang}, \binits{Y.}},
\oauthor{\bsnm{Shao}, \binits{J.}},
\oauthor{\bsnm{Song}, \binits{Y.}},
\oauthor{\bsnm{Wang}, \binits{Z.}},
\oauthor{\bsnm{Yuan}, \binits{C.}},
\oauthor{\bsnm{Zhang}, \binits{C.}},
\oauthor{\bsnm{Zhang}, \binits{H.}},
\oauthor{\bsnm{Zhuang}, \binits{W.}},
\oauthor{\bsnm{Li}, \binits{X.}}:
Ai flow: Perspectives, scenarios, and approaches.
arXiv preprint arXiv:2506.12479
(2025)
\end{botherref}
\endbibitem

\bibitem[\protect\citeauthoryear{Liu et~al.}{2025}]{liu2025fast}
\begin{barticle}
\bauthor{\bsnm{Liu}, \binits{D.}},
\bauthor{\bsnm{Wang}, \binits{Z.}},
\bauthor{\bsnm{Meng}, \binits{X.}}:
\batitle{Fast intensive crowd counting model of internet of things based on multi-scale attention mechanism}.
\bjtitle{IET Image Processing}
\bvolume{19}(\bissue{1}),
\bfpage{12686}
(\byear{2025})
\end{barticle}
\endbibitem

\bibitem[\protect\citeauthoryear{Li et~al.}{2018}]{CSRNet}
\begin{bchapter}
\bauthor{\bsnm{Li}, \binits{Y.}},
\bauthor{\bsnm{Zhang}, \binits{X.}},
\bauthor{\bsnm{Chen}, \binits{D.}}:
\bctitle{C{SRN}et: {D}ilated {C}onvolutional {N}eural {N}etworks for {U}nderstanding the {H}ighly {C}ongested {S}cenes}.
In: \bbtitle{Proc. IEEE Conference on Computer Vision and Pattern Recognition,},
pp. \bfpage{1091}--\blpage{1100}
(\byear{2018})
\end{bchapter}
\endbibitem

\bibitem[\protect\citeauthoryear{Idrees et~al.}{2018}]{UCF-QNRF}
\begin{bchapter}
\bauthor{\bsnm{Idrees}, \binits{H.}},
\bauthor{\bsnm{Tayyab}, \binits{M.}},
\bauthor{\bsnm{Athrey}, \binits{K.}},
\bauthor{\bsnm{Zhang}, \binits{D.}},
\bauthor{\bsnm{Al{-}M{\'{a}}adeed}, \binits{S.}},
\bauthor{\bsnm{Rajpoot}, \binits{N.}},
\bauthor{\bsnm{Shah}, \binits{M.}}:
\bctitle{Composition {L}oss for {C}ounting, {D}ensity {M}ap {E}stimation and {L}ocalization in {D}ense {C}rowds}.
In: \bbtitle{Proc. IEEE European Conference on Computer Vision},
vol. \bseriesno{11206},
pp. \bfpage{544}--\blpage{559}
(\byear{2018})
\end{bchapter}
\endbibitem

\bibitem[\protect\citeauthoryear{Wang et~al.}{2020}]{MobileCount}
\begin{barticle}
\bauthor{\bsnm{Wang}, \binits{P.}},
\bauthor{\bsnm{Gao}, \binits{C.}},
\bauthor{\bsnm{Wang}, \binits{Y.}},
\bauthor{\bsnm{Li}, \binits{H.}},
\bauthor{\bsnm{Gao}, \binits{Y.}}:
\batitle{Mobile{C}ount: {A}n {E}fficient {E}ncoder-{D}ecoder {F}ramework for {R}eal-{T}ime {C}rowd {C}ounting}.
\bjtitle{Neurocomputing}
\bvolume{407},
\bfpage{292}--\blpage{299}
(\byear{2020})
\end{barticle}
\endbibitem

\bibitem[\protect\citeauthoryear{Li}{2024}]{PN}
\begin{barticle}
\bauthor{\bsnm{Li}, \binits{X.}}:
\batitle{Positive-incentive noise}.
\bjtitle{IEEE Transactions on Neural Networks and Learning Systems}
\bvolume{35}(\bissue{6}),
\bfpage{8708}--\blpage{8714}
(\byear{2024})
\end{barticle}
\endbibitem

\bibitem[\protect\citeauthoryear{Gao et~al.}{2020}]{PCC-Net}
\begin{barticle}
\bauthor{\bsnm{Gao}, \binits{J.}},
\bauthor{\bsnm{Wang}, \binits{Q.}},
\bauthor{\bsnm{Li}, \binits{X.}}:
\batitle{{PCC-N}et: {P}erspective {C}rowd {C}ounting via {S}patial {C}onvolutional {N}etwork}.
\bjtitle{IEEE Transactions on Circuits and Systems for Video Technology}
\bvolume{30}(\bissue{10}),
\bfpage{3486}--\blpage{3498}
(\byear{2020})
\end{barticle}
\endbibitem

\bibitem[\protect\citeauthoryear{Liu et~al.}{2020}]{SKT}
\begin{bchapter}
\bauthor{\bsnm{Liu}, \binits{L.}},
\bauthor{\bsnm{Chen}, \binits{J.}},
\bauthor{\bsnm{Wu}, \binits{H.}},
\bauthor{\bsnm{Chen}, \binits{T.}},
\bauthor{\bsnm{Li}, \binits{G.}},
\bauthor{\bsnm{Lin}, \binits{L.}}:
\bctitle{Efficient {C}rowd {C}ounting via {S}tructured {K}nowledge {T}ransfer}.
In: \bbtitle{{MM} '20: The 28th {ACM} International Conference on Multimedia, Virtual Event / Seattle},
pp. \bfpage{2645}--\blpage{2654}
(\byear{2020})
\end{bchapter}
\endbibitem

\bibitem[\protect\citeauthoryear{Jiang et~al.}{2025}]{MiN}
\begin{botherref}
\oauthor{\bsnm{Jiang}, \binits{K.}},
\oauthor{\bsnm{Shi}, \binits{Z.}},
\oauthor{\bsnm{Zhang}, \binits{D.}},
\oauthor{\bsnm{Zhang}, \binits{H.}},
\oauthor{\bsnm{Li}, \binits{X.}}:
Mixture of noise for pre-trained model-based class-incremental learning.
arXiv preprint arXiv:2509.16738
(2025)
\end{botherref}
\endbibitem

\bibitem[\protect\citeauthoryear{Yu et~al.}{2021}]{Lite-HRNet}
\begin{bchapter}
\bauthor{\bsnm{Yu}, \binits{C.}},
\bauthor{\bsnm{Xiao}, \binits{B.}},
\bauthor{\bsnm{Gao}, \binits{C.}},
\bauthor{\bsnm{Yuan}, \binits{L.}},
\bauthor{\bsnm{Zhang}, \binits{L.}},
\bauthor{\bsnm{Sang}, \binits{N.}},
\bauthor{\bsnm{Wang}, \binits{J.}}:
\bctitle{Lite-{HRN}et: {A} {L}ightweight {H}igh-{R}esolution {N}etwork}.
In: \bbtitle{Proc. IEEE Conference on Computer Vision and Pattern Recognition},
pp. \bfpage{10440}--\blpage{10450}
(\byear{2021})
\end{bchapter}
\endbibitem

\bibitem[\protect\citeauthoryear{Zhang et~al.}{2024}]{PiNDA}
\begin{botherref}
\oauthor{\bsnm{Zhang}, \binits{H.}},
\oauthor{\bsnm{Xu}, \binits{Y.}},
\oauthor{\bsnm{Huang}, \binits{S.}},
\oauthor{\bsnm{Li}, \binits{X.}}:
Data augmentation of contrastive learning is estimating positive-incentive noise.
arXiv preprint arXiv:2408.09929
(2024)
\end{botherref}
\endbibitem

\bibitem[\protect\citeauthoryear{Huang et~al.}{2025}]{PiNGDA}
\begin{bchapter}
\bauthor{\bsnm{Huang}, \binits{S.}},
\bauthor{\bsnm{Xu}, \binits{Y.}},
\bauthor{\bsnm{Zhang}, \binits{H.}},
\bauthor{\bsnm{Li}, \binits{X.}}:
\bctitle{Learn beneficial noise as graph augmentation}.
In: \bbtitle{Proceedings of the 42nd International Conference on Machine Learning (ICML)}
(\byear{2025})
\end{bchapter}
\endbibitem

\bibitem[\protect\citeauthoryear{Lin et~al.}{2017}]{FPN}
\begin{bchapter}
\bauthor{\bsnm{Lin}, \binits{T.}},
\bauthor{\bsnm{Doll{\'{a}}r}, \binits{P.}},
\bauthor{\bsnm{Ross}, \binits{B.}},
\bauthor{\bsnm{He}, \binits{K.}},
\bauthor{\bsnm{Hariharan}, \binits{B.}},
\bauthor{\bsnm{Serge}, \binits{J.}}:
\bctitle{Feature {P}yramid {N}etworks for {O}bject {D}etection}.
In: \bbtitle{Proc. IEEE Conference on Computer Vision and Pattern Recognition},
pp. \bfpage{936}--\blpage{944}
(\byear{2017})
\end{bchapter}
\endbibitem

\bibitem[\protect\citeauthoryear{Li et~al.}{2017}]{group}
\begin{bchapter}
\bauthor{\bsnm{Li}, \binits{X.}},
\bauthor{\bsnm{Chen}, \binits{M.}},
\bauthor{\bsnm{Nie}, \binits{F.}},
\bauthor{\bsnm{Wang}, \binits{Q.}}:
\bctitle{A {M}ultiview-{B}ased {P}arameter {F}ree {F}ramework for {G}roup {D}etection}.
In: \bbtitle{Proc. AAAI Conference on Artificial Intelligence},
pp. \bfpage{4147}--\blpage{4153}
(\byear{2017})
\end{bchapter}
\endbibitem

\bibitem[\protect\citeauthoryear{Li et~al.}{2020}]{Quantifying}
\begin{barticle}
\bauthor{\bsnm{Li}, \binits{X.}},
\bauthor{\bsnm{Chen}, \binits{M.}},
\bauthor{\bsnm{Wang}, \binits{Q.}}:
\batitle{Quantifying and {D}etecting {C}ollective {M}otion in {C}rowd {S}cenes}.
\bjtitle{IEEE Transactions on Image Processing}
\bvolume{PP}(\bissue{99}),
\bfpage{1}--\blpage{1}
(\byear{2020})
\end{barticle}
\endbibitem

\bibitem[\protect\citeauthoryear{Zhang et~al.}{2018}]{zhang2018review}
\begin{barticle}
\bauthor{\bsnm{Zhang}, \binits{D.}},
\bauthor{\bsnm{Fu}, \binits{H.}},
\bauthor{\bsnm{Han}, \binits{J.}},
\bauthor{\bsnm{Borji}, \binits{A.}},
\bauthor{\bsnm{Li}, \binits{X.}}:
\batitle{A review of co-saliency detection algorithms: Fundamentals, applications, and challenges}.
\bjtitle{ACM Transactions on Intelligent Systems and Technology (TIST)}
\bvolume{9}(\bissue{4}),
\bfpage{1}--\blpage{31}
(\byear{2018})
\end{barticle}
\endbibitem

\bibitem[\protect\citeauthoryear{Viola and Jones}{2001}]{detection1}
\begin{bchapter}
\bauthor{\bsnm{Viola}, \binits{P.}},
\bauthor{\bsnm{Jones}, \binits{M.}}:
\bctitle{Robust real-time face detection}.
In: \bbtitle{Proc. IEEE International Conference on Computer Vision},
vol. \bseriesno{2},
pp. \bfpage{747}--\blpage{747}
(\byear{2001}).
\doiurl{10.1109/ICCV.2001.937709}
\end{bchapter}
\endbibitem

\bibitem[\protect\citeauthoryear{Felzenszwalb et~al.}{2010}]{detection2}
\begin{barticle}
\bauthor{\bsnm{Felzenszwalb}, \binits{P.}},
\bauthor{\bsnm{Girshick}, \binits{R.}},
\bauthor{\bsnm{McAllester}, \binits{D.}},
\bauthor{\bsnm{Ramanan}, \binits{D.}}:
\batitle{Object {D}etection with {D}iscriminatively {T}rained {P}art-{B}ased {M}odels}.
\bjtitle{IEEE Transactions on Pattern Analysis and Machine Intelligence}
\bvolume{32}(\bissue{9}),
\bfpage{1627}--\blpage{1645}
(\byear{2010})
\end{barticle}
\endbibitem

\bibitem[\protect\citeauthoryear{Han et~al.}{2022}]{han2022dr}
\begin{bchapter}
\bauthor{\bsnm{Han}, \binits{T.}},
\bauthor{\bsnm{Bai}, \binits{L.}},
\bauthor{\bsnm{Gao}, \binits{J.}},
\bauthor{\bsnm{Wang}, \binits{Q.}},
\bauthor{\bsnm{Ouyang}, \binits{W.}}:
\bctitle{Dr. vic: Decomposition and reasoning for video individual counting}.
In: \bbtitle{Proceedings of the IEEE/CVF Conference on Computer Vision and Pattern Recognition},
pp. \bfpage{3083}--\blpage{3092}
(\byear{2022})
\end{bchapter}
\endbibitem

\bibitem[\protect\citeauthoryear{Chan and Vasconcelos}{2009}]{regress}
\begin{bchapter}
\bauthor{\bsnm{Chan}, \binits{A.}},
\bauthor{\bsnm{Vasconcelos}, \binits{N.}}:
\bctitle{Bayesian {P}oisson {R}egression for {C}rowd {C}ounting},
pp. \bfpage{545}--\blpage{551}
(\byear{2009})
\end{bchapter}
\endbibitem

\bibitem[\protect\citeauthoryear{Idrees et~al.}{2013}]{regress1}
\begin{bchapter}
\bauthor{\bsnm{Idrees}, \binits{H.}},
\bauthor{\bsnm{Saleemi}, \binits{I.}},
\bauthor{\bsnm{Seibert}, \binits{C.}},
\bauthor{\bsnm{Shah}, \binits{M.}}:
\bctitle{Multi-{S}ource {M}ulti-{S}cale {C}ounting in {E}xtremely {D}ense {C}rowd {I}mages}.
In: \bbtitle{Proc. IEEE Conference on Computer Vision and Pattern Recognition},
pp. \bfpage{2547}--\blpage{2554}
(\byear{2013})
\end{bchapter}
\endbibitem

\bibitem[\protect\citeauthoryear{Wan et~al.}{2023}]{wan2023modeling}
\begin{barticle}
\bauthor{\bsnm{Wan}, \binits{J.}},
\bauthor{\bsnm{Wu}, \binits{Q.}},
\bauthor{\bsnm{Chan}, \binits{A.B.}}:
\batitle{Modeling noisy annotations for point-wise supervision}.
\bjtitle{IEEE Transactions on Pattern Analysis and Machine Intelligence}
\bvolume{45}(\bissue{12}),
\bfpage{15065}--\blpage{15080}
(\byear{2023})
\end{barticle}
\endbibitem

\bibitem[\protect\citeauthoryear{Wan et~al.}{2020}]{wan2020kernel}
\begin{barticle}
\bauthor{\bsnm{Wan}, \binits{J.}},
\bauthor{\bsnm{Wang}, \binits{Q.}},
\bauthor{\bsnm{Chan}, \binits{A.B.}}:
\batitle{Kernel-based density map generation for dense object counting}.
\bjtitle{IEEE Transactions on Pattern Analysis and Machine Intelligence}
\bvolume{44}(\bissue{3}),
\bfpage{1357}--\blpage{1370}
(\byear{2020})
\end{barticle}
\endbibitem

\bibitem[\protect\citeauthoryear{Shu et~al.}{2024}]{wan2024gene}
\begin{barticle}
\bauthor{\bsnm{Shu}, \binits{W.}},
\bauthor{\bsnm{Wan}, \binits{J.}},
\bauthor{\bsnm{Chan}, \binits{A.B.}}:
\batitle{Generalized characteristic function loss for crowd analysis in the frequency domain}.
\bjtitle{IEEE Transactions on Pattern Analysis and Machine Intelligence}
\bvolume{46}(\bissue{5}),
\bfpage{2882}--\blpage{2899}
(\byear{2024})
\doiurl{10.1109/TPAMI.2023.3336196}
\end{barticle}
\endbibitem

\bibitem[\protect\citeauthoryear{Xiong et~al.}{2017}]{xiong2017spatiotemporal}
\begin{bchapter}
\bauthor{\bsnm{Xiong}, \binits{F.}},
\bauthor{\bsnm{Shi}, \binits{X.}},
\bauthor{\bsnm{Yeung}, \binits{D.-Y.}}:
\bctitle{Spatiotemporal modeling for crowd counting in videos}.
In: \bbtitle{Proceedings of the IEEE International Conference on Computer Vision},
pp. \bfpage{5151}--\blpage{5159}
(\byear{2017})
\end{bchapter}
\endbibitem

\bibitem[\protect\citeauthoryear{Lin and Chan}{2024}]{lin2024fixed}
\begin{bchapter}
\bauthor{\bsnm{Lin}, \binits{W.}},
\bauthor{\bsnm{Chan}, \binits{A.B.}}:
\bctitle{A fixed-point approach to unified prompt-based counting}.
In: \bbtitle{Proceedings of the AAAI Conference on Artificial Intelligence},
vol. \bseriesno{38},
pp. \bfpage{3468}--\blpage{3476}
(\byear{2024})
\end{bchapter}
\endbibitem

\bibitem[\protect\citeauthoryear{Zhao et~al.}{2025}]{zhao2025density}
\begin{botherref}
\oauthor{\bsnm{Zhao}, \binits{C.}},
\oauthor{\bsnm{Wan}, \binits{J.}},
\oauthor{\bsnm{Chan}, \binits{A.B.}}:
Density-based object detection in crowded scenes.
arXiv preprint arXiv:2504.09819
(2025)
\end{botherref}
\endbibitem

\bibitem[\protect\citeauthoryear{Lin and Chan}{2023}]{lin2023optimal}
\begin{bchapter}
\bauthor{\bsnm{Lin}, \binits{W.}},
\bauthor{\bsnm{Chan}, \binits{A.B.}}:
\bctitle{Optimal transport minimization: Crowd localization on density maps for semi-supervised counting}.
In: \bbtitle{Proceedings of the IEEE/CVF Conference on Computer Vision and Pattern Recognition},
pp. \bfpage{21663}--\blpage{21673}
(\byear{2023})
\end{bchapter}
\endbibitem

\bibitem[\protect\citeauthoryear{Zhang et~al.}{2016}]{MCNN}
\begin{bchapter}
\bauthor{\bsnm{Zhang}, \binits{Y.}},
\bauthor{\bsnm{Zhou}, \binits{D.}},
\bauthor{\bsnm{Chen}, \binits{S.}},
\bauthor{\bsnm{Gao}, \binits{S.}},
\bauthor{\bsnm{Ma}, \binits{Y.}}:
\bctitle{Single-{I}mage {C}rowd {C}ounting via {M}ulti-{C}olumn {C}onvolutional {N}eural {N}etwork}.
In: \bbtitle{Proc. IEEE Conference on Computer Vision and Pattern Recognition},
pp. \bfpage{589}--\blpage{597}
(\byear{2016})
\end{bchapter}
\endbibitem

\bibitem[\protect\citeauthoryear{Yi et~al.}{2021}]{SACCN}
\begin{botherref}
\oauthor{\bsnm{Yi}, \binits{Q.}},
\oauthor{\bsnm{Liu}, \binits{Y.}},
\oauthor{\bsnm{Jiang}, \binits{A.}},
\oauthor{\bsnm{Li}, \binits{J.}},
\oauthor{\bsnm{Mei}, \binits{K.}},
\oauthor{\bsnm{Wang}, \binits{M.}}:
Scale-{A}ware {N}etwork with {R}egional and {S}emantic {A}ttentions for {C}rowd {C}ounting under {C}luttered {B}ackground.
CoRR
\textbf{abs/2101.01479}
(2021)
\end{botherref}
\endbibitem

\bibitem[\protect\citeauthoryear{Shi et~al.}{2020}]{C-CNN}
\begin{bchapter}
\bauthor{\bsnm{Shi}, \binits{X.}},
\bauthor{\bsnm{Li}, \binits{X.}},
\bauthor{\bsnm{Wu}, \binits{C.}},
\bauthor{\bsnm{Kong}, \binits{S.}},
\bauthor{\bsnm{Yang}, \binits{J.}},
\bauthor{\bsnm{He}, \binits{L.}}:
\bctitle{A {R}eal-{T}ime {D}eep {N}etwork for {C}rowd {C}ounting}.
In: \bbtitle{IEEE International Conference on Acoustics, Speech and Signal Processing},
pp. \bfpage{2328}--\blpage{2332}
(\byear{2020})
\end{bchapter}
\endbibitem

\bibitem[\protect\citeauthoryear{Yuan et~al.}{2025}]{yuan2025distance}
\begin{barticle}
\bauthor{\bsnm{Yuan}, \binits{Y.}},
\bauthor{\bsnm{Guo}, \binits{H.}},
\bauthor{\bsnm{Gao}, \binits{J.}}:
\batitle{Distance-aware network for physical-world object distribution estimation and counting}.
\bjtitle{Pattern Recognition}
\bvolume{157},
\bfpage{110896}
(\byear{2025})
\end{barticle}
\endbibitem

\bibitem[\protect\citeauthoryear{Lin et~al.}{2025}]{lin2025webuibench}
\begin{botherref}
\oauthor{\bsnm{Lin}, \binits{Z.}},
\oauthor{\bsnm{Zhou}, \binits{Z.}},
\oauthor{\bsnm{Zhao}, \binits{Z.}},
\oauthor{\bsnm{Wan}, \binits{T.}},
\oauthor{\bsnm{Ma}, \binits{Y.}},
\oauthor{\bsnm{Gao}, \binits{J.}},
\oauthor{\bsnm{Li}, \binits{X.}}:
Webuibench: A comprehensive benchmark for evaluating multimodal large language models in webui-to-code.
arXiv preprint arXiv:2506.07818
(2025)
\end{botherref}
\endbibitem

\bibitem[\protect\citeauthoryear{Li et~al.}{2025}]{li2025llms}
\begin{botherref}
\oauthor{\bsnm{Li}, \binits{H.}},
\oauthor{\bsnm{Gao}, \binits{H.}},
\oauthor{\bsnm{Zhao}, \binits{Z.}},
\oauthor{\bsnm{Lin}, \binits{Z.}},
\oauthor{\bsnm{Gao}, \binits{J.}},
\oauthor{\bsnm{Li}, \binits{X.}}:
Llms caught in the crossfire: Malware requests and jailbreak challenges.
arXiv preprint arXiv:2506.10022
(2025)
\end{botherref}
\endbibitem

\bibitem[\protect\citeauthoryear{Huang et~al.}{2025}]{PiNI}
\begin{bchapter}
\bauthor{\bsnm{Huang}, \binits{S.}},
\bauthor{\bsnm{Zhang}, \binits{H.}},
\bauthor{\bsnm{Li}, \binits{X.}}:
\bctitle{Enhance vision-language alignment with noise}.
In: \bbtitle{Proceedings of the AAAI Conference on Artificial Intelligence},
vol. \bseriesno{39},
pp. \bfpage{17449}--\blpage{17457}
(\byear{2025})
\end{bchapter}
\endbibitem

\bibitem[\protect\citeauthoryear{Wan et~al.}{2024}]{wan2024robust}
\begin{bchapter}
\bauthor{\bsnm{Wan}, \binits{J.}},
\bauthor{\bsnm{Wu}, \binits{Q.}},
\bauthor{\bsnm{Lin}, \binits{W.}},
\bauthor{\bsnm{Chan}, \binits{A.}}:
\bctitle{Robust zero-shot crowd counting and localization with adaptive resolution sam}.
In: \bbtitle{European Conference on Computer Vision},
pp. \bfpage{478}--\blpage{495}
(\byear{2024}).
\bcomment{Springer}
\end{bchapter}
\endbibitem

\bibitem[\protect\citeauthoryear{Fan et~al.}{2024}]{fan2024learning}
\begin{botherref}
\oauthor{\bsnm{Fan}, \binits{Y.}},
\oauthor{\bsnm{Wan}, \binits{J.}},
\oauthor{\bsnm{Ma}, \binits{A.J.}}:
Learning crowd scale and distribution for weakly supervised crowd counting and localization.
IEEE Transactions on Circuits and Systems for Video Technology
(2024)
\end{botherref}
\endbibitem

\bibitem[\protect\citeauthoryear{Wang et~al.}{2025}]{wang2025diffusion}
\begin{bchapter}
\bauthor{\bsnm{Wang}, \binits{Z.}},
\bauthor{\bsnm{Li}, \binits{Y.}},
\bauthor{\bsnm{Wan}, \binits{J.}},
\bauthor{\bsnm{Vasconcelos}, \binits{N.}}:
\bctitle{Diffusion-based data augmentation for object counting problems}.
In: \bbtitle{ICASSP 2025-2025 IEEE International Conference on Acoustics, Speech and Signal Processing (ICASSP)},
pp. \bfpage{1}--\blpage{5}
(\byear{2025}).
\bcomment{IEEE}
\end{bchapter}
\endbibitem

\bibitem[\protect\citeauthoryear{Bai et~al.}{2024}]{bai2024combating}
\begin{bchapter}
\bauthor{\bsnm{Bai}, \binits{S.}},
\bauthor{\bsnm{Li}, \binits{S.}},
\bauthor{\bsnm{Zhuang}, \binits{W.}},
\bauthor{\bsnm{Zhang}, \binits{J.}},
\bauthor{\bsnm{Yang}, \binits{K.}},
\bauthor{\bsnm{Hou}, \binits{J.}},
\bauthor{\bsnm{Yi}, \binits{S.}},
\bauthor{\bsnm{Zhang}, \binits{S.}},
\bauthor{\bsnm{Gao}, \binits{J.}}:
\bctitle{Combating data imbalances in federated semi-supervised learning with dual regulators}.
In: \bbtitle{Proceedings of the AAAI Conference on Artificial Intelligence},
vol. \bseriesno{38},
pp. \bfpage{10989}--\blpage{10997}
(\byear{2024})
\end{bchapter}
\endbibitem

\bibitem[\protect\citeauthoryear{Han et~al.}{2015}]{han2015two}
\begin{barticle}
\bauthor{\bsnm{Han}, \binits{J.}},
\bauthor{\bsnm{Zhang}, \binits{D.}},
\bauthor{\bsnm{Wen}, \binits{S.}},
\bauthor{\bsnm{Guo}, \binits{L.}},
\bauthor{\bsnm{Liu}, \binits{T.}},
\bauthor{\bsnm{Li}, \binits{X.}}:
\batitle{Two-stage learning to predict human eye fixations via sdaes}.
\bjtitle{IEEE transactions on cybernetics}
\bvolume{46}(\bissue{2}),
\bfpage{487}--\blpage{498}
(\byear{2015})
\end{barticle}
\endbibitem

\bibitem[\protect\citeauthoryear{Tao et~al.}{2008}]{tao2008bayesian}
\begin{barticle}
\bauthor{\bsnm{Tao}, \binits{D.}},
\bauthor{\bsnm{Song}, \binits{M.}},
\bauthor{\bsnm{Li}, \binits{X.}},
\bauthor{\bsnm{Shen}, \binits{J.}},
\bauthor{\bsnm{Sun}, \binits{J.}},
\bauthor{\bsnm{Wu}, \binits{X.}},
\bauthor{\bsnm{Faloutsos}, \binits{C.}},
\bauthor{\bsnm{Maybank}, \binits{S.J.}}:
\batitle{Bayesian tensor approach for 3-d face modeling}.
\bjtitle{IEEE Transactions on Circuits and Systems for Video Technology}
\bvolume{18}(\bissue{10}),
\bfpage{1397}--\blpage{1410}
(\byear{2008})
\end{barticle}
\endbibitem

\bibitem[\protect\citeauthoryear{Zhao et~al.}{2021}]{zhao2021reconstructive}
\begin{barticle}
\bauthor{\bsnm{Zhao}, \binits{B.}},
\bauthor{\bsnm{Li}, \binits{H.}},
\bauthor{\bsnm{Lu}, \binits{X.}},
\bauthor{\bsnm{Li}, \binits{X.}}:
\batitle{Reconstructive sequence-graph network for video summarization}.
\bjtitle{IEEE Transactions on Pattern Analysis and Machine Intelligence}
\bvolume{44}(\bissue{5}),
\bfpage{2793}--\blpage{2801}
(\byear{2021})
\end{barticle}
\endbibitem

\bibitem[\protect\citeauthoryear{N. et~al.}{2016}]{SqueezeNet}
\begin{botherref}
\oauthor{\bsnm{N.}, \binits{F.}},
\oauthor{\bsnm{Han}, \binits{S.}},
\oauthor{\bsnm{W.}, \binits{M.}},
\oauthor{\bsnm{Ashraf}, \binits{K.}},
\oauthor{\bsnm{J.}, \binits{W.}},
\oauthor{\bsnm{Keutzer}, \binits{K.}}:
Squeeze{N}et: {A}lex{N}et-level {A}ccuracy with 50{X} {F}ewer {P}arameters and <0.5{MB} {M}odel {S}ize
(2016)
\end{botherref}
\endbibitem

\bibitem[\protect\citeauthoryear{Zhou et~al.}{2024}]{zhou2024rrtrn}
\begin{barticle}
\bauthor{\bsnm{Zhou}, \binits{Q.}},
\bauthor{\bsnm{Gao}, \binits{J.}},
\bauthor{\bsnm{Yuan}, \binits{Y.}},
\bauthor{\bsnm{Wang}, \binits{Q.}}:
\batitle{Rrtrn: A lightweight and effective backbone for scene text recognition}.
\bjtitle{Expert Systems with Applications}
\bvolume{243},
\bfpage{122769}
(\byear{2024})
\end{barticle}
\endbibitem

\bibitem[\protect\citeauthoryear{G. et~al.}{2017}]{MobileNets}
\begin{botherref}
\oauthor{\bsnm{G.}, \binits{A.}},
\oauthor{\bsnm{Zhu}, \binits{M.}},
\oauthor{\bsnm{Chen}, \binits{B.}},
\oauthor{\bsnm{Kalenichenko}, \binits{D.}},
\oauthor{\bsnm{Wang}, \binits{W.}},
\oauthor{\bsnm{Weyand}, \binits{T.}},
\oauthor{\bsnm{Andreetto}, \binits{M.}},
\oauthor{\bsnm{Adam}, \binits{H.}}:
Mobile{N}ets: {E}fficient {C}onvolutional {N}eural {N}etworks for {M}obile {V}ision {A}pplications.
CoRR
\textbf{abs/1704.04861}
(2017)
\end{botherref}
\endbibitem

\bibitem[\protect\citeauthoryear{Zhang et~al.}{2018}]{ShuffleNet}
\begin{bchapter}
\bauthor{\bsnm{Zhang}, \binits{X.}},
\bauthor{\bsnm{Zhou}, \binits{X.}},
\bauthor{\bsnm{Lin}, \binits{M.}},
\bauthor{\bsnm{Sun}, \binits{J.}}:
\bctitle{Shuffle{N}et: {A}n {E}xtremely {E}fficient {C}onvolutional {N}eural {N}etwork for {M}obile {D}evices}.
In: \bbtitle{Proc. IEEE Conference on Computer Vision and Pattern Recognition},
pp. \bfpage{6848}--\blpage{6856}
(\byear{2018})
\end{bchapter}
\endbibitem

\bibitem[\protect\citeauthoryear{Sandler et~al.}{2018}]{MobileNetV2}
\begin{bchapter}
\bauthor{\bsnm{Sandler}, \binits{M.}},
\bauthor{\bsnm{Andrew}, \binits{G.}},
\bauthor{\bsnm{Zhu}, \binits{M.}},
\bauthor{\bsnm{Zhmoginov}, \binits{A.}},
\bauthor{\bsnm{Chen}, \binits{L.}}:
\bctitle{Mobile{N}et{V}2: {I}nverted {R}esiduals and {L}inear {B}ottlenecks}.
In: \bbtitle{Proc. IEEE Conference on Computer Vision and Pattern Recognition},
pp. \bfpage{4510}--\blpage{4520}
(\byear{2018})
\end{bchapter}
\endbibitem

\bibitem[\protect\citeauthoryear{Ma et~al.}{2018}]{ShuffleNetV2}
\begin{botherref}
\oauthor{\bsnm{Ma}, \binits{N.}},
\oauthor{\bsnm{Zhang}, \binits{X.}},
\oauthor{\bsnm{Zheng}, \binits{H.-T.}},
\oauthor{\bsnm{Sun}, \binits{J.}}:
Shuffle{N}et{V}2: {P}ractical {G}uidelines for {E}fficient {C}NN {A}rchitecture {D}esign
(2018)
\end{botherref}
\endbibitem

\bibitem[\protect\citeauthoryear{Ding et~al.}{2024}]{ding2024ff}
\begin{botherref}
\oauthor{\bsnm{Ding}, \binits{H.}},
\oauthor{\bsnm{Gao}, \binits{J.}},
\oauthor{\bsnm{Yuan}, \binits{Y.}},
\oauthor{\bsnm{Wang}, \binits{Q.}}:
Ff-lpd: A real-time frame-by-frame license plate detector with knowledge distillation and feature propagation.
IEEE Transactions on Image Processing
(2024)
\end{botherref}
\endbibitem

\bibitem[\protect\citeauthoryear{Zhou et~al.}{2025}]{zhou2025scale}
\begin{botherref}
\oauthor{\bsnm{Zhou}, \binits{Q.}},
\oauthor{\bsnm{Gao}, \binits{J.}},
\oauthor{\bsnm{Wang}, \binits{Q.}}:
Scale efficient training for large datasets.
IEEE/CVF Conference on Computer Vision and Pattern Recognition
(2025)
\end{botherref}
\endbibitem

\bibitem[\protect\citeauthoryear{Guo et~al.}{2025}]{guo2025enhancing}
\begin{bchapter}
\bauthor{\bsnm{Guo}, \binits{H.}},
\bauthor{\bsnm{Gao}, \binits{J.}},
\bauthor{\bsnm{Yuan}, \binits{Y.}}:
\bctitle{Enhancing low-rank adaptation with recoverability-based reinforcement pruning for object counting}.
In: \bbtitle{Proceedings of the AAAI Conference on Artificial Intelligence},
vol. \bseriesno{39},
pp. \bfpage{3238}--\blpage{3246}
(\byear{2025})
\end{bchapter}
\endbibitem

\bibitem[\protect\citeauthoryear{Tao et~al.}{2008}]{FaceModeling}
\begin{barticle}
\bauthor{\bsnm{Tao}, \binits{D.}},
\bauthor{\bsnm{Song}, \binits{M.}},
\bauthor{\bsnm{Li}, \binits{X.}},
\bauthor{\bsnm{Shen}, \binits{J.}},
\bauthor{\bsnm{Sun}, \binits{J.}},
\bauthor{\bsnm{Wu}, \binits{X.}},
\bauthor{\bsnm{Faloutsos}, \binits{C.}},
\bauthor{\bsnm{Maybank}, \binits{S.J.}}:
\batitle{Bayesian tensor approach for 3-d face modeling}.
\bjtitle{IEEE Transactions on Circuits and Systems for Video Technology}
\bvolume{18}(\bissue{10}),
\bfpage{1397}--\blpage{1410}
(\byear{2008})
\end{barticle}
\endbibitem

\bibitem[\protect\citeauthoryear{Zhang et~al.}{2024}]{zhang2024integrating}
\begin{botherref}
\oauthor{\bsnm{Zhang}, \binits{D.}},
\oauthor{\bsnm{Wang}, \binits{F.}},
\oauthor{\bsnm{Ning}, \binits{L.}},
\oauthor{\bsnm{Zhao}, \binits{Z.}},
\oauthor{\bsnm{Gao}, \binits{J.}},
\oauthor{\bsnm{Li}, \binits{X.}}:
Integrating sam with feature interaction for remote sensing change detection.
IEEE Transactions on Geoscience and Remote Sensing
(2024)
\end{botherref}
\endbibitem

\bibitem[\protect\citeauthoryear{Wang et~al.}{2024}]{wang2024embedding}
\begin{botherref}
\oauthor{\bsnm{Wang}, \binits{Q.}},
\oauthor{\bsnm{Jia}, \binits{Y.}},
\oauthor{\bsnm{Gao}, \binits{J.}},
\oauthor{\bsnm{Li}, \binits{Q.}}:
Embedding generalized semantic knowledge into few-shot remote sensing segmentation.
IEEE Transactions on Geoscience and Remote Sensing
(2024)
\end{botherref}
\endbibitem

\bibitem[\protect\citeauthoryear{Gao et~al.}{2025}]{gao2025combining}
\begin{botherref}
\oauthor{\bsnm{Gao}, \binits{J.}},
\oauthor{\bsnm{Zhang}, \binits{D.}},
\oauthor{\bsnm{Wang}, \binits{F.}},
\oauthor{\bsnm{Ning}, \binits{L.}},
\oauthor{\bsnm{Zhao}, \binits{Z.}},
\oauthor{\bsnm{Li}, \binits{X.}}:
Combining sam with limited data for change detection in remote sensing.
IEEE Transactions on Geoscience and Remote Sensing
(2025)
\end{botherref}
\endbibitem

\bibitem[\protect\citeauthoryear{Han et~al.}{2015}]{EyeFixations}
\begin{barticle}
\bauthor{\bsnm{Han}, \binits{J.}},
\bauthor{\bsnm{Zhang}, \binits{D.}},
\bauthor{\bsnm{Wen}, \binits{S.}},
\bauthor{\bsnm{Guo}, \binits{L.}},
\bauthor{\bsnm{Liu}, \binits{T.}},
\bauthor{\bsnm{Li}, \binits{X.}}:
\batitle{Two-stage learning to predict human eye fixations via sdaes}.
\bjtitle{IEEE transactions on cybernetics}
\bvolume{46}(\bissue{2}),
\bfpage{487}--\blpage{498}
(\byear{2015})
\end{barticle}
\endbibitem

\bibitem[\protect\citeauthoryear{Zhang et~al.}{2025}]{VPN}
\begin{botherref}
\oauthor{\bsnm{Zhang}, \binits{H.}},
\oauthor{\bsnm{Huang}, \binits{S.}},
\oauthor{\bsnm{Guo}, \binits{Y.}},
\oauthor{\bsnm{Li}, \binits{X.}}:
Variational positive-incentive noise: How noise benefits models.
IEEE Transactions on Pattern Analysis and Machine Intelligence
(2025)
\end{botherref}
\endbibitem

\bibitem[\protect\citeauthoryear{Cao et~al.}{2018}]{SANet}
\begin{bchapter}
\bauthor{\bsnm{Cao}, \binits{X.}},
\bauthor{\bsnm{Wang}, \binits{Z.}},
\bauthor{\bsnm{Zhao}, \binits{Y.}},
\bauthor{\bsnm{Su}, \binits{F.}}:
\bctitle{Scale {A}ggregation {N}etwork for {A}ccurate and {E}fficient {C}rowd {C}ounting}.
In: \bbtitle{Proc. IEEE European Conference on Computer Vision},
vol. \bseriesno{11209},
pp. \bfpage{757}--\blpage{773}
(\byear{2018})
\end{bchapter}
\endbibitem

\bibitem[\protect\citeauthoryear{Vishwanath and Vishal}{2017}]{CMTL}
\begin{bchapter}
\bauthor{\bsnm{Vishwanath}, \binits{A.}},
\bauthor{\bsnm{Vishal}, \binits{M.}}:
\bctitle{C{NN}-{B}ased {C}ascaded {M}ulti-{T}ask {L}earning of {H}igh-{L}evel {P}rior and {D}ensity {E}stimation for {C}rowd {C}ounting}.
In: \bbtitle{Proc. IEEE International Conference on Advanced Video and Signal Based Surveillance},
pp. \bfpage{1}--\blpage{6}
(\byear{2017})
\end{bchapter}
\endbibitem

\bibitem[\protect\citeauthoryear{Shen et~al.}{2018}]{ACSCP}
\begin{bchapter}
\bauthor{\bsnm{Shen}, \binits{Z.}},
\bauthor{\bsnm{Xu}, \binits{Y.}},
\bauthor{\bsnm{Ni}, \binits{B.}},
\bauthor{\bsnm{Wang}, \binits{M.}},
\bauthor{\bsnm{Hu}, \binits{J.}},
\bauthor{\bsnm{Yang}, \binits{X.}}:
\bctitle{Crowd {C}ounting via {A}dversarial {C}ross-{S}cale {C}onsistency {P}ursuit}.
In: \bbtitle{Proc. IEEE Conference on Computer Vision and Pattern Recognition},
pp. \bfpage{5245}--\blpage{5254}
(\byear{2018}).
\doiurl{10.1109/CVPR.2018.00550}
\end{bchapter}
\endbibitem

\bibitem[\protect\citeauthoryear{Sam and Babu}{2018}]{TDF-CNN}
\begin{bchapter}
\bauthor{\bsnm{Sam}, \binits{D.B.}},
\bauthor{\bsnm{Babu}, \binits{R.V.}}:
\bctitle{Top-{D}own {F}eedback for {C}rowd {C}ounting {C}onvolutional {N}eural {N}etwork}.
In: \bbtitle{Proc. AAAI Conference on Artificial Intelligence},
pp. \bfpage{7323}--\blpage{7330}
(\byear{2018})
\end{bchapter}
\endbibitem

\bibitem[\protect\citeauthoryear{Liu et~al.}{2022}]{Liu2022}
\begin{barticle}
\bauthor{\bsnm{Liu}, \binits{Y.}},
\bauthor{\bsnm{Cao}, \binits{G.}},
\bauthor{\bsnm{Shi}, \binits{H.}},
\bauthor{\bsnm{Hu}, \binits{Y.}}:
\batitle{Lw-{C}ount: {A}n {E}ffective {L}ightweight {E}ncoding-{D}ecoding {C}rowd {C}ounting {N}etwork}.
\bjtitle{IEEE Transactions on Circuits and Systems for Video Technology}
\bvolume{32}(\bissue{10}),
\bfpage{6821}--\blpage{6834}
(\byear{2022})
\end{barticle}
\endbibitem

\bibitem[\protect\citeauthoryear{Gao et~al.}{2019}]{C3F-VGG}
\begin{botherref}
\oauthor{\bsnm{Gao}, \binits{J.}},
\oauthor{\bsnm{Lin}, \binits{W.}},
\oauthor{\bsnm{Zhao}, \binits{B.}},
\oauthor{\bsnm{Wang}, \binits{D.}},
\oauthor{\bsnm{Gao}, \binits{C.}},
\oauthor{\bsnm{Wen}, \binits{J.}}:
C{\^{}}3 {F}ramework: {A}n {O}pen-{S}ource {P}y{T}orch {C}ode for {C}rowd {C}ounting.
CoRR
\textbf{abs/1907.02724}
(2019)
\end{botherref}
\endbibitem

\bibitem[\protect\citeauthoryear{Wang et~al.}{2019}]{SFCN+}
\begin{bchapter}
\bauthor{\bsnm{Wang}, \binits{Q.}},
\bauthor{\bsnm{Gao}, \binits{J.}},
\bauthor{\bsnm{Lin}, \binits{W.}},
\bauthor{\bsnm{Yuan}, \binits{Y.}}:
\bctitle{Learning from {S}ynthetic {D}ata for {C}rowd {C}ounting in the {W}ild}.
In: \bbtitle{Proc. IEEE Conference on Computer Vision and Pattern Recognition},
pp. \bfpage{8198}--\blpage{8207}
(\byear{2019})
\end{bchapter}
\endbibitem

\bibitem[\protect\citeauthoryear{Wang et~al.}{2024}]{wang2024rt3c}
\begin{botherref}
\oauthor{\bsnm{Wang}, \binits{R.}},
\oauthor{\bsnm{Hao}, \binits{Y.}},
\oauthor{\bsnm{Miao}, \binits{Y.}},
\oauthor{\bsnm{Hu}, \binits{L.}},
\oauthor{\bsnm{Chen}, \binits{M.}}:
Rt3c: Real-time crowd counting in multi-scene video streams via cloud-edge-device collaboration.
IEEE Transactions on Services Computing
(2024)
\end{botherref}
\endbibitem

\bibitem[\protect\citeauthoryear{Sun et~al.}{2024}]{sun2024strmt}
\begin{barticle}
\bauthor{\bsnm{Sun}, \binits{L.}},
\bauthor{\bsnm{Zhao}, \binits{J.}},
\bauthor{\bsnm{Zhang}, \binits{J.}},
\bauthor{\bsnm{Zhang}, \binits{F.}},
\bauthor{\bsnm{Ye}, \binits{K.}},
\bauthor{\bsnm{Xu}, \binits{C.}}:
\batitle{Strmt: A state transition based model for real-time crowd counting in a metro system}.
\bjtitle{Concurrency and Computation: Practice and Experience}
\bvolume{36}(\bissue{14}),
\bfpage{8086}
(\byear{2024})
\end{barticle}
\endbibitem

\bibitem[\protect\citeauthoryear{Wang et~al.}{2021}]{NWPU-Crowd}
\begin{barticle}
\bauthor{\bsnm{Wang}, \binits{Q.}},
\bauthor{\bsnm{Gao}, \binits{J.}},
\bauthor{\bsnm{Lin}, \binits{W.}},
\bauthor{\bsnm{Li}, \binits{X.}}:
\batitle{N{WPU}-{C}rowd: {A} {L}arge-{S}cale {B}enchmark for {C}rowd {C}ounting and {L}ocalization}.
\bjtitle{IEEE Transactions on Pattern Analysis and Machine Intelligence}
\bvolume{43}(\bissue{6}),
\bfpage{2141}--\blpage{2149}
(\byear{2021})
\end{barticle}
\endbibitem

\end{thebibliography}

\end{document}